\documentclass{article}
\usepackage{arxiv}

\usepackage{amssymb}
\usepackage{amsmath}
\usepackage{graphicx}
\usepackage{booktabs}
\usepackage{siunitx}
\usepackage{longtable,tabularx}
\usepackage{subcaption}
\usepackage{caption}
\usepackage{float}
\usepackage{array}
\usepackage{multirow}
\usepackage{pifont}
\usepackage{adjustbox}
\usepackage{tikz}
\usepackage[colorlinks=true,linkcolor=blue,citecolor=blue]{hyperref}
\usepackage[numbers,sort&compress]{natbib}
\newcommand{\cmark}{\ding{51}}
\newcommand{\xmark}{\ding{55}}

\hypersetup{
  pdfauthor={},
  pdftitle={BlendedNet++: A dataset and benchmark for field-resolved aerodynamics and inverse design of blended wing body aircraft.},
  pdfsubject={},
  pdfkeywords={}
}

\title{BlendedNet++: A dataset and benchmark for field-resolved aerodynamics and inverse design of blended wing body aircraft.}

\author{
  Nicholas Sung\thanks{These authors contributed equally to this work.} \\
  Department of Mechanical Engineering \\
  Massachusetts Institute of Technology \\
  Cambridge, MA \\
  \texttt{nicksung@mit.edu}
  \And
  Steven Spreizer\footnotemark[1] \\
  MIT Lincoln Laboratory \\
  Lexington, MA \\
  \texttt{steven.spreizer@ll.mit.edu}
  \And
  Mohamed Elrefaie \\
  Department of Mechanical Engineering \\
  Massachusetts Institute of Technology \\
  Cambridge, MA \\
  \texttt{mohamed.elrefaie@mit.edu}
  \And
  Matthew C.\ Jones \\
  MIT Lincoln Laboratory \\
  Lexington, MA \\
  \texttt{matthew.jones@ll.mit.edu}
  \And
  Faez Ahmed \\
  Department of Mechanical Engineering \\
  Massachusetts Institute of Technology \\
  Cambridge, MA \\
  \texttt{faez@mit.edu}
}

\begin{document}
\maketitle

\begin{abstract}
The conceptual design of Blended Wing Body (BWB) aircraft is often constrained by the high computational cost of resolving complex aerodynamics over a high-dimensional design space. While deep learning offers a pathway to rapid aerodynamic prediction and inverse design, its adoption in aerospace engineering is limited by a lack of large-scale, field-resolved training data. This work addresses this gap by introducing \textbf{BlendedNet++}, a comprehensive aerodynamic dataset comprising 12,492 unique BWB geometries, each evaluated using steady Reynolds-Averaged Navier--Stokes (RANS) simulations to provide integrated forces and dense surface fields ($C_p, C_f$). Leveraging this data, we establish a robust framework for two critical engineering tasks: (1) real-time prediction of surface aerodynamic fields using geometric deep learning models, and (2) generative inverse design. We benchmark five surrogate architectures, identifying Transolver as the most accurate for  field predictions. Furthermore, we demonstrate a generative inverse design pipeline using conditional diffusion models combined with gradient-based refinement. This hybrid approach is shown to generate multiple feasible designs that satisfy specific lift-to-drag targets with high accuracy ($R^2 > 0.99$), as confirmed by computational fluid dynamics (CFD) simulation. These resources enable a shift from iterative analysis to direct generation in early-stage BWB design.
\end{abstract}


\section*{Nomenclature}
\noindent(All symbols include units where applicable; ``--'' denotes dimensionless.)

{\small
\setlength{\LTleft}{0pt}
\setlength{\LTright}{0pt}
\renewcommand\arraystretch{1.05}
\begin{longtable}{@{}l@{\quad=\quad}>{\raggedright\arraybackslash}p{0.78\linewidth}@{}}

\multicolumn{2}{@{}l}{\textbf{Aerodynamic Coefficients}}\\
$C_L$ & lift coefficient [--] \\
$C_D$ & drag coefficient [--] \\
$C_M$ & pitching-moment coefficient about nose reference [--] \\
$C_p$ & surface pressure coefficient [--] \\
$C_{f_x}$ & skin-friction coefficient, streamwise component [--] \\
$C_{f_y}$ & skin-friction coefficient, spanwise component [--] \\
$C_{f_z}$ & skin-friction coefficient, vertical component [--] \\
$L/D$ & lift-to-drag ratio [--] \\
\addlinespace[3pt]

\multicolumn{2}{@{}l}{\textbf{Geometry and Coordinates}}\\
$\mathbf{x}=(x,y,z)$ & Cartesian coordinates on surface mesh [m] \\
$\mathbf{n}=(n_x,n_y,n_z)$ & outward unit surface normal [--] \\
$C_1$ & centerline length (reference length) [m] \\
$C_2,\,C_3,\,C_4$ & chord lengths at planform stations [m] \\
$C_2/C_1,\,C_3/C_1,\,C_4/C_1$ & chord-length ratios (station-to-centerline) [--] \\
$B_1,\,B_2,\,B_3$ & spanwise station locations measured from centerline [m] \\
$B_1/C_1,\,B_2/C_1,\,B_3/C_1$ & span-fraction parameters [--] \\
$X_3$ & streamwise location of outboard break [m] \\
$S_1,\,S_3$ & inboard / outboard leading-edge sweep angles [deg] \\
$y^+$ & non-dimensional wall distance [--] \\
\addlinespace[3pt]

\multicolumn{2}{@{}l}{\textbf{Flight Conditions}}\\
$\mathrm{alt}_{\mathrm{kft}}$ & altitude [kft] \\
$M_\infty$ & free-stream Mach number [--] \\
$Re_L$ & Reynolds number based on $L=C_1$ [--] \\
$\alpha$ & angle of attack [deg] \\
\addlinespace[3pt]

\multicolumn{2}{@{}l}{\textbf{Learning / Inference Variables}}\\
$\mathbf{p}\in\mathbb{R}^9$ &
planform parameter vector:
$(B_1/C_1,B_2/C_1,B_3/C_1,\,C_2/C_1,\,C_3/C_1,\,C_4/C_1,\,S_1,\,S_3,\,X_3/C_1)$ \\
$\boldsymbol{\mu}$ &
conditioning vector for inverse design:
$(\mathrm{alt}_{\mathrm{kft}},\log_{10} Re_L,M_\infty,\alpha,(L/D)_{\mathrm{target}})$ \\
$a_t, \bar{a}_t$ & diffusion process coefficients [--] \\
$\theta$ & conditional diffusion model neural network parameters [--] \\

\end{longtable}}

\section{Introduction}
The Blended Wing Body (BWB) configuration represents a promising generational leap in aircraft design, offering significant improvements in aerodynamic efficiency and fuel economy compared to conventional tube-and-wing architectures. By integrating the fuselage and wings into a single lifting surface, BWBs reduce wetted area and allow structural loads to align more naturally with aerodynamic lift \cite{liebeck2004design}. Early feasibility studies and recent assessments have corroborated these advantages, reporting potential fuel burn reductions of 20--30\% alongside substantial structural and acoustic benefits \cite{carter2009blended, sax40_shielding, zhenli2019assessment}. 

However, realizing these gains requires navigating a design space that is far more sensitive to local flow physics than conventional aircraft. High-fidelity Computational Fluid Dynamics (CFD) is necessary to capture these effects but remains too computationally intensive for the large-scale design sweeps required during conceptual design. ~\cite{HighFidelityCFD,MultiFidelityBWB}.

\textbf{The Necessity of High Fidelity Field-Resolved Data in BWB Design.}
Traditional conceptual design for aircraft typically relies on integrated force coefficients ($C_L, C_D, C_M$) and low-fidelity simulations like the Vortex Lattice Method (VLM). However, these tools are often insufficient for assessing the true multidisciplinary viability of a highly integrated airframe like the Blended-Wing-Body (BWB). Designs that appear optimal under lower-fidelity analysis frequently prove unviable when subjected to high-fidelity modeling, which can reveal critical issues such as insufficient static margins or localized aerodynamic loads that necessitate excessive structural reinforcement \cite{yazdi2026optimization}. Consequently, field-resolved surface data, specifically pressure ($C_p$) and skin-friction ($C_f$) distributions, are essential early in the design cycle for two primary reasons:

First, because the BWB is a tailless configuration, its longitudinal stability and control are governed entirely by the local pressure behavior over the airframe's trailing edges. Minor errors in predicting flow separation can lead to significant discrepancies in the pitching moment, resulting in substantial ``trim penalties'' that can erase the aircraft's predicted efficiency gains if not accounted for during initial sizing \cite{GrayZingg2024, RoysdonKhalid2012}.

Second, the BWB centerbody must function as a non-cylindrical pressure vessel. Unlike the efficient membrane stress (hoop tension) found in a standard cylindrical fuselage, the flat or shallow-curved sections of a BWB fuselage must resist internal cabin pressure through bending. Because these bending stresses are an order of magnitude higher than membrane stresses, accurate $C_p$ mapping is required to quantify external suction, which serves as a vital load-alleviating factor against internal pressure \cite{NASA_BWB_Fuselage, liebeck2004design}. Neglecting these granular distributions can lead to structural weight estimate errors that challenge the configuration's economic feasibility \cite{bradley2004sizing}.

\textbf{Why machine learning?} Data-driven surrogates can approximate high-fidelity solvers at orders-of-magnitude lower runtime, enabling rapid screening, sensitivity analysis, and inverse design under tight iteration budgets~\cite{dong2021deep,sabater2022fast}. Yet surrogates trained on datasets with highly restricted design space bounds or insufficient parametric variance often struggle to generalize to unseen geometries or novel flight conditions. This challenge is amplified for \emph{field-level} prediction, where learning dense surface quantities ($C_p$, $C_f$) is high-dimensional and requires many \emph{unique} shapes with consistent labels to capture local boundary-layer behavior and nonlocal interactions across the airframe. Empirically, recent field-resolved works in external aerodynamics (e.g., \emph{DrivAerNet++}) show that scaling dataset size enables the model to capture more complex flow physics, ultimately improving the real-world applicability and out-of-distribution robustness of the predictions~\cite{elrefaie2024drivaernet++, elrefaie2025drivaernet, elrefaie2024drivaernet}. Robust surrogates, therefore, require large, diverse, field-resolved datasets with standardized splits and metrics for fair assessment and reproducibility. Because architectures encode different inductive biases such as point set, graph, transformer/operator, and Feature-wise Linear Modulation (FiLM) style conditioning, robust conclusions require a standardized benchmark across model families. We evaluate five complementary families on a shared, geometry-disjoint split to compare accuracy and scalability on equal footing.

\textbf{BlendedNet\footnote{Dataset available at: \url{https://doi.org/10.7910/DVN/VJT9EP}}}~\cite{blendednet} took an initial step toward addressing data scarcity with 1099 geometries and dense surface outputs across multiple flight conditions, demonstrating that data-driven surrogates can predict pointwise pressure and skin-friction fields with low error.
Yet, broader exploration of BWB design still lacks (i) a densely sampled geometric design space, (ii) standardized, geometry-disjoint splits and consistent forward benchmarks across diverse machine learning model families, and (iii) a reproducible inverse-design methodological framework, providing a generative AI and optimization baseline beyond purely gradient-based surrogates.

\textbf{BlendedNet++\footnote{Dataset available at: \url{https://doi.org/10.7910/DVN/ICIDK4}}} expands upon the foundation of the original dataset: \emph{12{,}492} distinct BWB geometries at one flight condition each, all with integrated coefficients and dense surface fields.
We (1) document the generation of an updated BWB dataset prioritizing geometric variation, (2) release geometry-disjoint splits, (3) establish a five-model forward benchmark spanning graph, point-set, coordinate-transformer, Feature-wise Linear Modulation Network (FiLMNet), and operator-learning paradigms, and (4) compare a conditional diffusion model, a gradient-based optimizer, and a diffusion–optimization hybrid that first samples diverse candidates with the conditional diffusion model and then locally optimizes them for near-exact target lift to drag adherence.z

 Our contributions are the following:
 \begin{itemize}
    \item \textbf{Scaled BWB dataset with surface fields.} 12{,}492 geometries (one flight condition each) with $C_L$, $C_D$, $C_M$ and dense surface $C_p$, $(C_{f_x},C_{f_y},C_{f_z})$, enabling field-level learning at scale.
    \item \textbf{Tactical Unmanned Aircraft System (UAS) benchmarks.} Geometry-disjoint train-test split, featuring a standardized test set specifically  for Group 3 UAS flight envelopes ($\mathrm{alt} \le \qty{18}{kft}$) to evaluate model performance on realistic tactical mission profiles.
    \item \textbf{Five-family forward benchmark.} FiLMNet, Transolver (a coordinate Transformer), PointNet, GraphUNet, and a General Neural Operator Transformer (GNOT) with unified metrics on pointwise fields.
    \item \textbf{Inverse-design baseline.} We compare a conditional diffusion model (conditioned on flight conditions and target $L/D$), a gradient-based optimizer on the same surrogate and box constraints, and a \emph{diffusion–optimization hybrid} that first samples diverse candidates with the conditional diffusion model and then locally refines them through the same gradient-based optimizer for near-exact target lift to drag adherence.
\end{itemize}

\begin{figure}[h!]
  \centering
  \includegraphics[width=\textwidth]{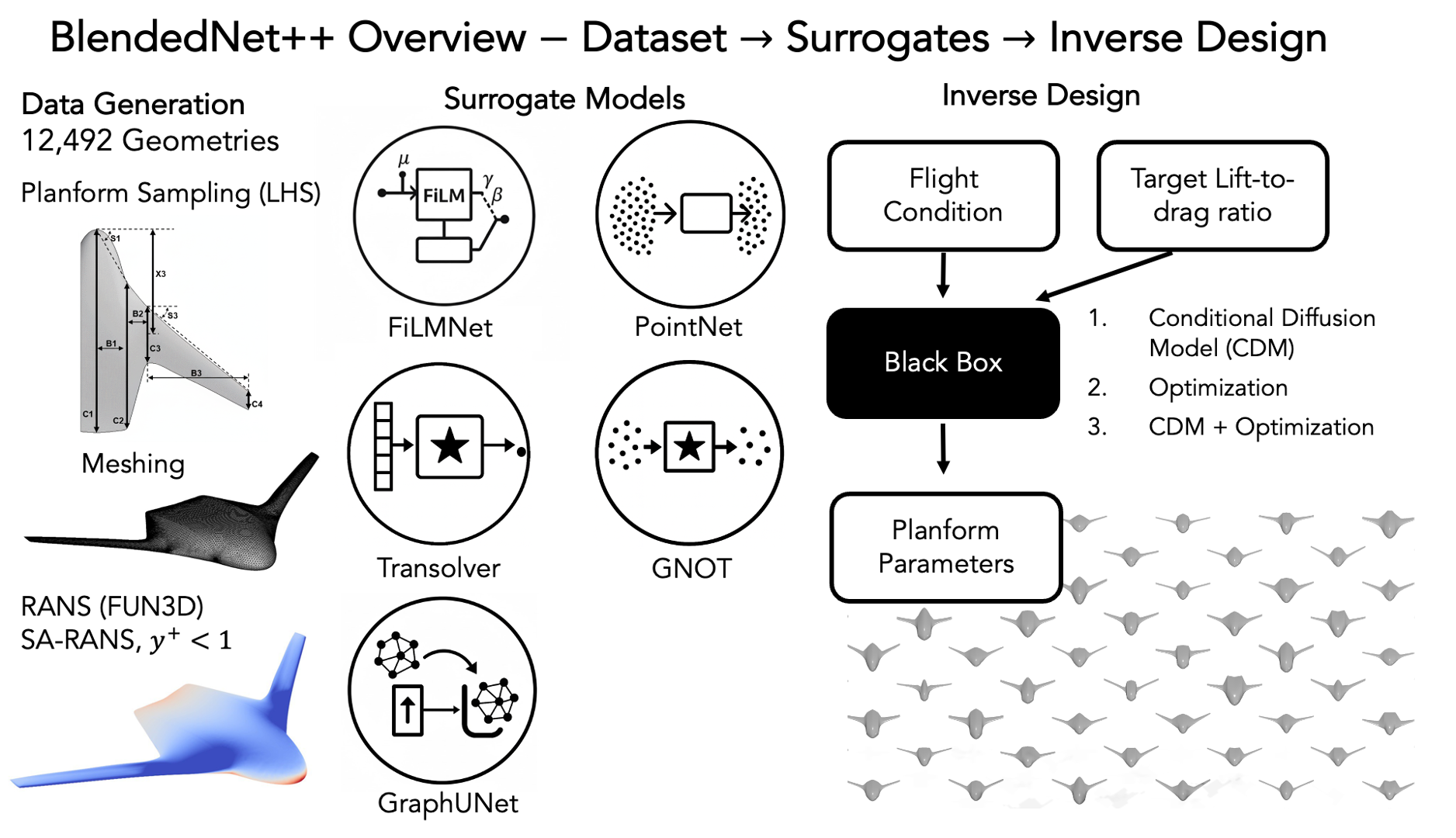}
  \caption{BlendedNet++ overview: (left) geometry parameterization and meshing,
  (middle) surrogate benchmarks for field-level prediction (FiLMNet, Transolver, PointNet, GraphUNet, GNOT), and (right) inverse-design pipeline (conditional diffusion, gradient-based optimization) over the 9-D planform box. The dataset couples integrated coefficients $(C_L,C_D,C_M)$ with dense surface fields $C_p$ and $C_f$ for 12{,}492 unique BWB geometries.}
  \label{fig:overview}
\end{figure}

Our objective is to facilitate systematic investigation of field-level aerodynamics, to enable rigorous and standardized model comparisons, and to provide a reproducible framework that supports future scaling and multi-condition dataset extensions. 
An overview of the data pipeline, surrogate families, and inverse-design setup is shown in Fig.~\ref{fig:overview}.

\begin{table}[h!]
    \centering
    \caption{Comparison of a few 3D Aircraft Aerodynamic Datasets and Case Studies}
    \label{tab:aero_datasets}
    \resizebox{\textwidth}{!}{ 
    \begin{tabular}{@{\hskip 6pt}l@{\hskip 6pt} l@{\hskip 6pt} c@{\hskip 6pt} c@{\hskip 6pt} c@{\hskip 6pt} c@{\hskip 6pt} c@{\hskip 6pt} c@{\hskip 6pt} c@{\hskip 6pt}} 
        \toprule
        \textbf{Type} & \textbf{Name} & \textbf{Size (geometries × flight conditions/geometry)} & \multicolumn{3}{c}{\textbf{Aerodynamic Values}} & \textbf{Modalities} & \textbf{BWB} & \textbf{Open Source} \\
        \cmidrule(lr){4-6}
        & & & $C_L$ / $C_D$ & $C_M$ & Pointwise $C_p$ / $C_f$ &  &  &  \\ 
        \midrule
        \multirow{4}{*}{Case Study} 
        & NASA BWB-450-1L \cite{carter2009blended} & 1 geometry, 27 conditions & \cmark & \cmark & \xmark & Physical & \cmark & \xmark \\
        & NASA/Boeing X-48B BWB \cite{liebeck2004design} & 2 geometries, 10+ conditions & \cmark & \cmark & \xmark & Physical & \cmark & \xmark \\
        & SACCON Dataset \cite{loeser2010saccon} & 1 geometry, 40K+ conditions & \cmark & \cmark & \xmark & Physical & \cmark & \xmark \\
        & NASA CRM \cite{rivers2019nasa} (DPW \cite{sclafani2013dpw} + ETW\cite{lutz2013going} CFD) & 1 geometry & \cmark & \cmark & \cmark & Mesh, Parametric & \xmark & Partial\textsuperscript{*} \\
        \midrule
        \multirow{4}{*}{Dataset} 
        & AircraftVerse\cite{cobb2023aircraftverse} & 27,714 geometries & \cmark & \xmark & \xmark & Mesh, Parametric & \xmark & \cmark \\
        & ShapeNet Aircraft\cite{edwards2021design} & 4,045 geometries & \cmark & \xmark & \xmark & Mesh, 2D images & \xmark & \cmark \\
        & ONERA CRM WBPN\cite{peter2025onera} & 1 geometry, 468 flight conditions & \cmark & \cmark & \cmark & Mesh, Parametric & \xmark & \cmark \\
        & AASM Benchmark Case 4 (NASA CRM)\cite{bekemeyer2025aasm} & 1 geometry, 149 flight conditions & \cmark & \cmark & \cmark & Mesh, Parametric & \xmark & \cmark \\
        & HiLiftAeroML\cite{ashton2026hiliftaeroml} & 180 geometries, 10 flight conditions each & \cmark & \cmark & \cmark & Mesh, Parametric & \cmark & \cmark \\
        & SHIFTWing \cite{shift_wing_2025} & 200 geometries, $\sim$10 flight conditions each & \cmark & \cmark & \cmark & Mesh, Parametric & \xmark & \cmark \\
        & Martín et al. (2025) \cite{martin2025generative} & 1,500 geometries, 1 condition & \cmark & \cmark & \xmark & Parametric & \cmark & \xmark \\
        & BlendedNet\cite{blendednet} & 1099 geometries, $\sim 10$ conditions  & \cmark & \cmark & \cmark & Mesh, Parametric & \cmark & \cmark \\
        & BlendedNet++ & 12,492 geometries, 1 condition  & \cmark & \cmark & \cmark & Mesh, Parametric & \cmark & \cmark \\
        \bottomrule
    \end{tabular}
    }
    \vspace{0.5em}
    \footnotesize{\textsuperscript{*}Select mesh files (`.bdf`) and structural results (`.f06`) are available, but full aerodynamic datasets (e.g., full CFD or wind tunnel data) are not openly accessible.}
\end{table}

\section{Related Work}
\subsection{BWB Design and Optimization Studies}
The BWB configuration has been a focal point of radical aircraft research since the 1990s. Seminal programs such as the NASA/Boeing X-48 flight demonstrators established the configuration's promise, while the \emph{Silent Aircraft Initiative} (SAX-40) demonstrated that BWBs could serve as acoustic shields for embedded engines, achieving significant community noise reduction without sacrificing fuel burn \cite{liebeck2004design, carter2009blended, sax40_shielding}. 

State-of-the-art BWB design has matured from simple parametric scouts to high-fidelity Multidisciplinary Design Optimization (MDO) frameworks \cite{okonkwo2016review, ali2025review}. Early European collaborative efforts, such as the Multidisciplinary Design and Optimization (MOB) project, initially highlighted the power of combining low-fidelity panel methods with high-fidelity RANS through variable-fidelity inverse design approaches to optimize spanwise twist distributions \cite{qin2004aerodynamic}. These studies established that BWB design is the ``science of compromise'' \cite{qin2004aerodynamic}, where operational viability requires that the cruise point, maximum efficiency ($L/D$), and pitch trim be aligned simultaneously at a single flight attitude \cite{li2012aerodynamic}. 

The Zingg group has further quantified this aerodynamic potential, proving that high-fidelity cruise $L/D$ ratios as high as 21.7 for regional-class aircraft \cite{GrayZingg2024} and 28.4 for twin-aisle configurations \cite{yazdi2026optimization} are achievable when stability and cabin layout constraints are simultaneously satisfied. However, a critical finding from these studies is that BWB performance is systematically overpredicted if off-design requirements—such as takeoff field length, rotation ability, and one-engine-inoperative (OEI) trim—are excluded \cite{GrayZingg2024, yazdi2026optimization}. Furthermore, recent NATO AVT-298 (SWIFT configuration) studies underscore that BWB physics are uniquely sensitive to Reynolds number effects, particularly at high-lift takeoff conditions where industry-standard low-Reynolds facilities fail to represent full-scale separation behavior \cite{coppin2025introduction}. These findings highlight a major computational bottleneck: while RANS-based MDO is necessary for viability, it is computationally prohibitive for the wide-ranging design sweeps needed to scout the BWB manifold. This motivates the development of \textbf{fast, field-resolved surrogates} that can approximate these high-fidelity sensitivities in real-time.

Furthermore, Propulsion-Airframe Integration (PAI) and the management of vortical structures remain central challenges. Concepts like the NASA N3-X rely on Boundary Layer Ingestion (BLI), where integrated nacelles induce complex mutual interference effects that alter the airframe's static margin and lift distribution \cite{kim2016n3x, Liou2017AerodynamicDO}. As Ali et al. summarize, the absence of traditional stabilizers requires high-fidelity analysis to resolve the dynamic coupling of longitudinal and lateral modes \cite{ali2025review}. Recent iterative evaluations of high-capacity BWB configurations for long-range missions have demonstrated that even minor geometry modifications based on field analysis can significantly shift the center of pressure and stability characteristics \cite{dehpanah2015aerodynamic}. This iterative cycle of evaluation and modification, as seen in Dehpanah and Nejat, directly underscores the value of our proposed \textbf{generative inverse design pipeline}, which can automate the discovery of aerodynamically sound airframes from target performance metrics. Additionally, BWB UAS platforms have emerged as cost-effective research configurations for validating these structural and aerodynamic integration methods \cite{zhang2025uav}.

\subsection*{Public Datasets and Benchmarks}

Existing 3D aerodynamic resources are generally split into high-fidelity case studies and large-scale geometric corpora, as summarized in Table~\ref{tab:aero_datasets}. Case studies, such as the NASA Common Research Model (CRM) \cite{rivers2019nasa} and the X-48B program \cite{liebeck2004design}, provide gold-standard reference data but are typically restricted to a single or very few geometries evaluated across multiple flight conditions. While invaluable for validation, their lack of geometric diversity limits their utility for training generalizable Machine Learning (ML) surrogates. Conversely, broad geometric corpora like AircraftVerse \cite{cobb2023aircraftverse} and ShapeNet-Aircraft \cite{edwards2021design} provide the thousands of unique shapes required for deep learning, however, they often lack critical aerodynamic labels, particularly the pitching-moment coefficient ($C_M$) and field-resolved surface quantities.

As shown in Table~\ref{tab:aero_datasets}, there is a distinct gap in open-source data that combines high-volume geometric variety with field-resolved labels ($C_p, C_f$). This gap is particularly acute for the BWB configuration. While recent generative studies by Mart{\'i}n et al. \cite{martin2025generative} have expanded the BWB design space to 1,500 geometries, their labels are limited to integrated forces. As established in the preceding analysis, integrated coefficients alone are insufficient for BWB conceptual design, as they cannot resolve the local pressure distributions necessary to calculate structural bending penalties or the precise trailing-edge flow physics required for longitudinal trim.

BlendedNet++ addresses this deficiency by scaling the field-resolved approach of the original BlendedNet \cite{blendednet} by an order of magnitude. With 12,492 unique geometries, it provides the statistical density required to treat BWB aerodynamics as a continuous field-prediction task.

\subsection*{Machine Learning for Aerodynamic Prediction}

Machine learning (ML) is increasingly employed to mitigate the high computational overhead and turnaround time associated with high-fidelity aerodynamic studies. While purely physics-based Computational Fluid Dynamics (CFD) is the standard for physical accuracy, the solvers can be computationally expensive and sensitive to numerical noise. This noise typically manifests as high-frequency oscillations in residuals or integrated force coefficients resulting from local mesh quality issues or discretization artifacts \cite{dong2021deep, sabater2022fast}. Data-driven surrogates, particularly Deep Neural Networks (DNNs), can act as a regularizer in this context; they learn to smooth these numerical non-linearities to provide a more stable and continuous response surface for optimization \cite{sabater2022fast}. 

In conventional aircraft settings, researchers have utilized Gaussian processes and artificial neural networks to model aerodynamic performance \cite{chen2021framework, secco2017artificial}. More advanced architectures, such as Convolutional Neural Networks (CNNs) and autoencoders, are used to handle higher-dimensional shapes, while Graph Neural Networks (GNNs) align naturally with mesh-based CFD data \cite{dong2021deep}. 

The task of predicting dense surface fields, specifically the pressure coefficient ($C_p$) and skin friction coefficient ($\mathbf{C}_f$), remains a challenge due to the lack of large-scale, high-fidelity datasets. Recent benchmarks like DrivAerNet++ \cite{elrefaie2024drivaernet++} have demonstrated that scaling both dataset size and physical fidelity is critical for improving predictive accuracy and capturing complex scaling behaviors. To address these needs, several model families have emerged for \emph{surface-field} prediction:
\begin{itemize}
  \item \textbf{Coordinate transformers (Transolver):} These tokenize geometry and physics features to enable high-resolution field prediction with near-linear complexity \cite{wu2024Transolver}.
  \item \textbf{Hypernetworks (FiLM-Net):} These inject operating-point context, such as Mach number and Angle of Attack (AoA), via feature-wise scale and shift modulation throughout the network architecture \cite{perez2018film}.
  \item \textbf{Point-set models (PointNet):} These operate directly on unordered surface point clouds using shared Multilayer Perceptrons (MLPs) and permutation-invariant pooling to handle irregular sampling without remeshing \cite{qi2017pointnet}.
  \item \textbf{Hierarchical graph encoders (Graph U-Net):} These use pooling and unpooling layers to combine local pressure gradients with global planform trends, improving multi-scale reasoning \cite{gao2019graphunet}.
  \item \textbf{Operator-style transformers on graphs (GNOT):} These incorporate attention mechanisms to pass physical information across distant regions of the aircraft surface \cite{hao2023gnot}.

\end{itemize}

These families differ mainly in how they represent geometry (points, graphs, or coordinates) and how they handle long-range interactions (message passing vs.\ attention).

\subsection*{Inverse Design and Generative Modeling}

\paragraph{Classical inverse design}
Inverse aerodynamic design has traditionally relied on gradient-based and global optimization. Gradient-based methods, particularly those utilizing adjoint equations, are highly effective when smooth design spaces and reliable initializations are available \cite{lyu2014bwb, martins2013mdo}. These methods prioritize exact constraint handling and \emph{CFD-consistent gradients}. This means that the sensitivity of the objective function (e.g., $L/D$) with respect to design variables derived from the surrogate model must closely match the sensitivities calculated by the high-fidelity solver. 

When design objectives are rugged or multimodal, global strategies like Bayesian optimization are preferred, often within surrogate-assisted loops \cite{feldstein2020multifidelity,sarkar2019multifidelity,charisi2025multi}. However, these approaches become sample-intensive when the feasible region is narrow or when designs are \emph{strongly coupled to flight conditions}. This coupling refers to multi-point design requirements where a single geometry must satisfy conflicting performance targets across different operating points (e.g., maximizing cruise efficiency while maintaining takeoff stall margins), necessitating a much denser sampling of the joint design-flight condition space.

\paragraph{Learning-based inverse design}
More recently, machine learning has been used to accelerate inverse design. Supervised regressors can map targeted performance to design variables (e.g., mission-level or geometric quantities for BWB-type aircraft \cite{sharma2024mission}), while differentiable surrogates enable gradient-based optimization instead of repeated CFD solves (see also mesh/graph/coordinate surrogates in \cite{dong2021deep,perez2018film,catalani2024neural}). Generative models broaden this toolbox by modeling design distributions conditioned on goals or operating points. In 2D airfoil settings, conditional diffusion has been shown to generate shapes that meet coefficient targets \cite{wei2024diffairfoil,graves2024airfoil}, and subsequent work extends conditional diffusion to flying-wing concepts across multiple flight conditions \cite{lin2025flyingwing}. Recently, Martín et al.\ applied conditional diffusion to the 3D BWB setting, training on a parametric dataset of 1,500 geometries and showing the ability to generate diverse, aerodynamically valid designs directly from target $C_L$, $C_D$, and $C_M$ \cite{martin2025generative}.
Compared to direct optimization, generative approaches naturally capture multi-modality in the inverse map (many designs achieving similar targets) and can be combined with downstream refinement or optimization \cite{Sung2024Cooling}. Across these trends, an emerging pattern is to pair fast learned surrogates (for objectives and constraints) with search or generation mechanisms such as gradient-based, Bayesian/global, or diffusion-based to navigate large, constrained aerodynamic design spaces under realistic flight conditioning.

\subsection*{Limitations of Prior Work}
Despite progress, publicly available BWB resources rarely provide \emph{field-resolved} surface data at scale. Most offer only integrated coefficients ($C_L$, $C_D$, $C_M$) or 2D/airfoil cases; when 3D data exist, they usually contain \emph{few geometries} evaluated across many flight conditions, which complicates geometry-dependent generalization. The work that follows targets these gaps.

\section{Dataset Generation}
The data generation process for this work follows the same process as BlendedNet~\cite{blendednet} with some changes to parameter bounds as well as improvements to the simulation workflow.
A blended wing body (BWB) model was parameterically varied to create a set of geometries which are used to create volume meshes for computational fluid dynamics (CFD).
The CFD simulations yield integrated force coefficients (1D data) as well as the distribution of pressure and skin friction coefficients across the surface of the aircraft (2D) data. 
Future work may also include the 3D flow fields, but these are excluded here to balance dataset size and ease of use/distribution.

\subsection*{Geometry Generation}
\begin{figure}[h!]
    \centering
    \includegraphics[width=0.3\linewidth]{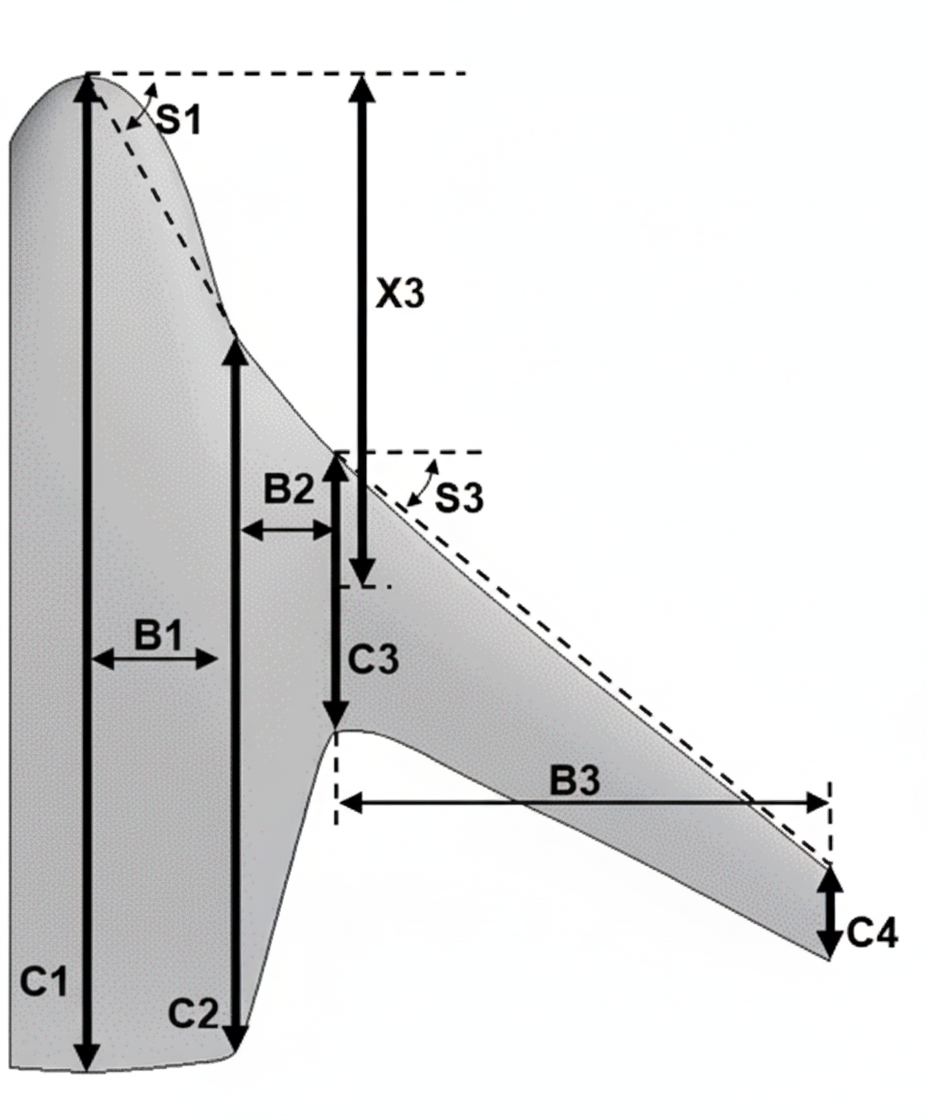}
    \caption{Parameterization of the BWB planform.}
    \label{fig:planform_parameters}
\end{figure}

\begin{figure}
    \centering
    \includegraphics[width=\linewidth]{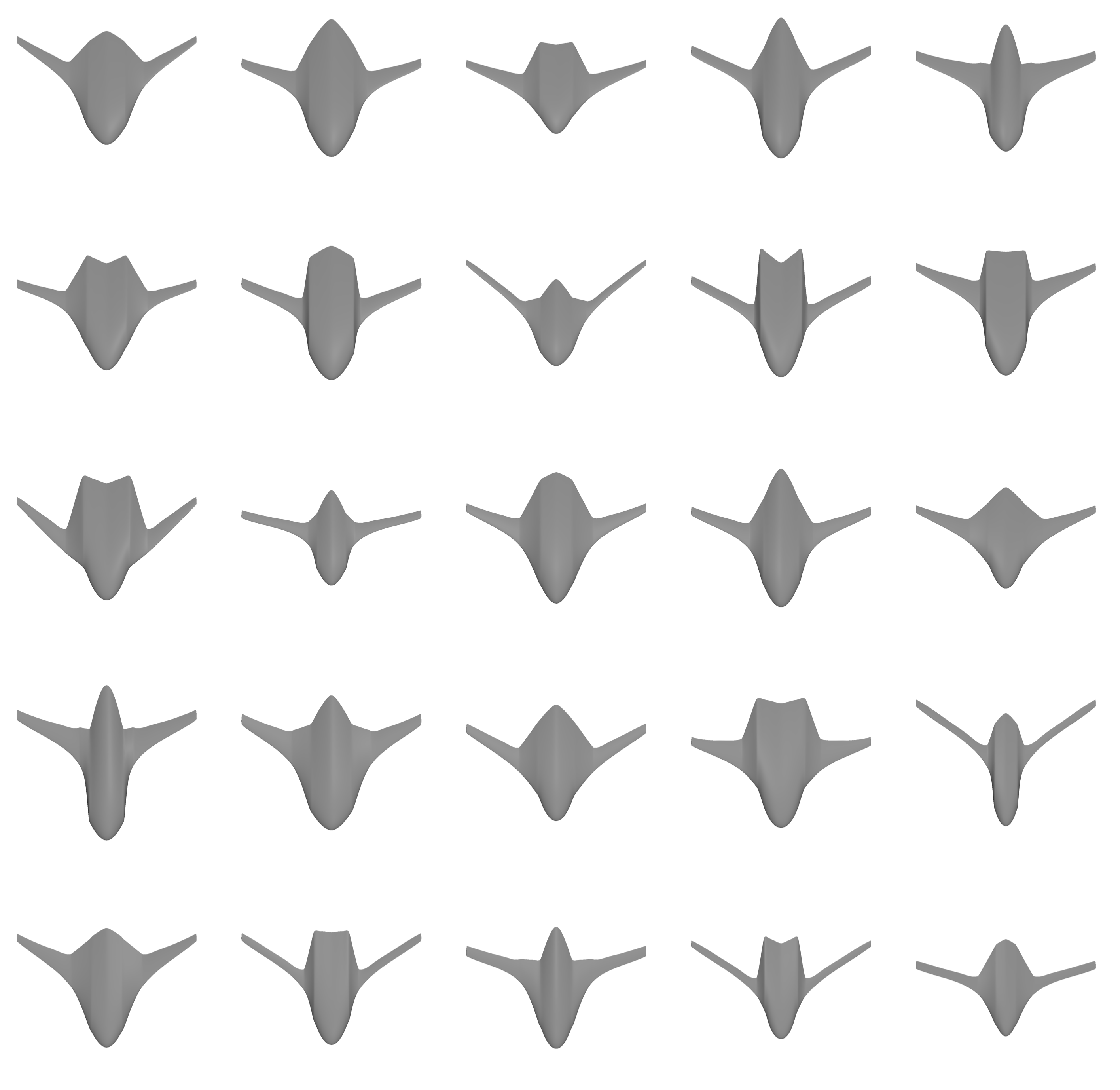}
    \caption{Representative renderings from BlendedNet++, illustrating the shape diversity across the dataset. The variations in fuselage geometry, wing span, and overall configuration highlight the richness of aerodynamic design space captured in the dataset.}
    \label{fig:airplane_diagonal}
\end{figure}

The parameterization of the BWB in this work is focused on planform variation with fixed airfoils for all the geometries.
Figure~\ref{fig:planform_parameters} illustrates the parameterization of the BWB planform. 
All length-based parameters are expressed as a ratio of the vehicle centerline length, $C_1$, to normalize lengths to the overall vehicle dimension.
When redimensionalized by the vehicle centerline length, this parameterization yields 3 chord lengths, 3 span distances, the streamwise location of the mid-chord of the wing root ($X_3$), and 2 sweep angles.
A summary of these parameters and their ranges can be found in Table~\ref{tab:param_bounds}.
The selected planform parameters and their ranges were inspired by Zhang et al.~\cite{zhang2016conceptual}.
The parameter ranges were adjusted slightly to broaden the range of generated planforms.
Table~\ref{tab:airfoils} lists the selected airfoil for each section, the selection of which were informed by Trac-Pho~\cite{trac-pho-masters} with the $C_4$ section as a symmetric airfoil for automatic CFD mesh generation to be more robust.
The outboard wing section has a 3 degree dihedral and no twist.
Figure~\ref{fig:airplane_diagonal} shows a sampling of generated geometries.
While some geometries are infeasible for the design of an actual aircraft (for example excessively slender wings), inclusion in the dataset still provides valuable data for surrogate models to learn from.

\begin{table}[h]
\centering
\caption{Blended wing body planform parameter bounds}
\label{tab:param_bounds}
\begin{tabular}{ccc}
\toprule
\textbf{Parameter} & \textbf{Min} & \textbf{Max} \\
\midrule
$C_2/C_1$ & 0.55 & 0.85 \\
$C_3/C_1$ & 0.18 & 0.28 \\
$C_4/C_1$ & 0.06 & 0.09 \\
$B_1/C_1$ & 0.10 & 0.20 \\
$B_2/C_1$ & 0.05 & 0.20 \\
$B_3/C_1$ & 0.35 & 0.70 \\
$X_3/C_1$ & 0.50 & 0.65 \\
$S_1$ [$^\circ$] & 40 & 60 \\
$S_3$ [$^\circ$] & 20 & 40 \\
\bottomrule
\end{tabular}
\end{table}

\begin{table}[h]
\centering
\caption{Blended wing body airfoil selections}
\label{tab:airfoils}
\begin{tabular}{cc}
\toprule
\textbf{Station} & \textbf{Airfoil} \\
\midrule
1 & NACA 25118 \\
2 & NACA 25118 \\
3 & NASA SC(2)-0412 \\
4 & NACA 0012 \\
\bottomrule
\end{tabular}
\end{table}

\begin{figure}
    \centering
    \includegraphics[width=\linewidth]{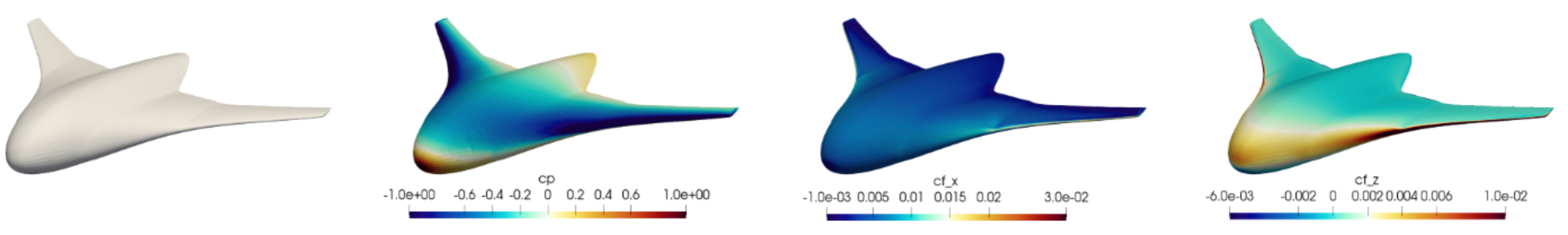}
    \caption{Example from BlendedNet++, showing the 3D shape alongside surface flow quantities. From left to right: baseline geometry, surface pressure coefficient (\(C_p\)), and skin-friction coefficients in the streamwise (\(C_{fx}\)), and vertical (\(C_{fz}\)) directions.}
    \label{fig:sample_case}
\end{figure}


The geometries are modeled using the Open Vehicle Sketch Pad (OpenVSP) geometric modeling software \cite{OpenVSP}.
Latin Hypercube Sampling (LHS) was used to select parameters to cover the design space of geometries.
For the training dataset 10k geometries were generated.
An additional 2.5k independently sampled geometries were created for the test set.
All geometries were exported as STEP (.stp) CAD files to feed into the meshing process.

\subsection*{Meshing}
\begin{figure}
    \centering
    \begin{subfigure}{0.4\textwidth} 
        \includegraphics[width=\textwidth]{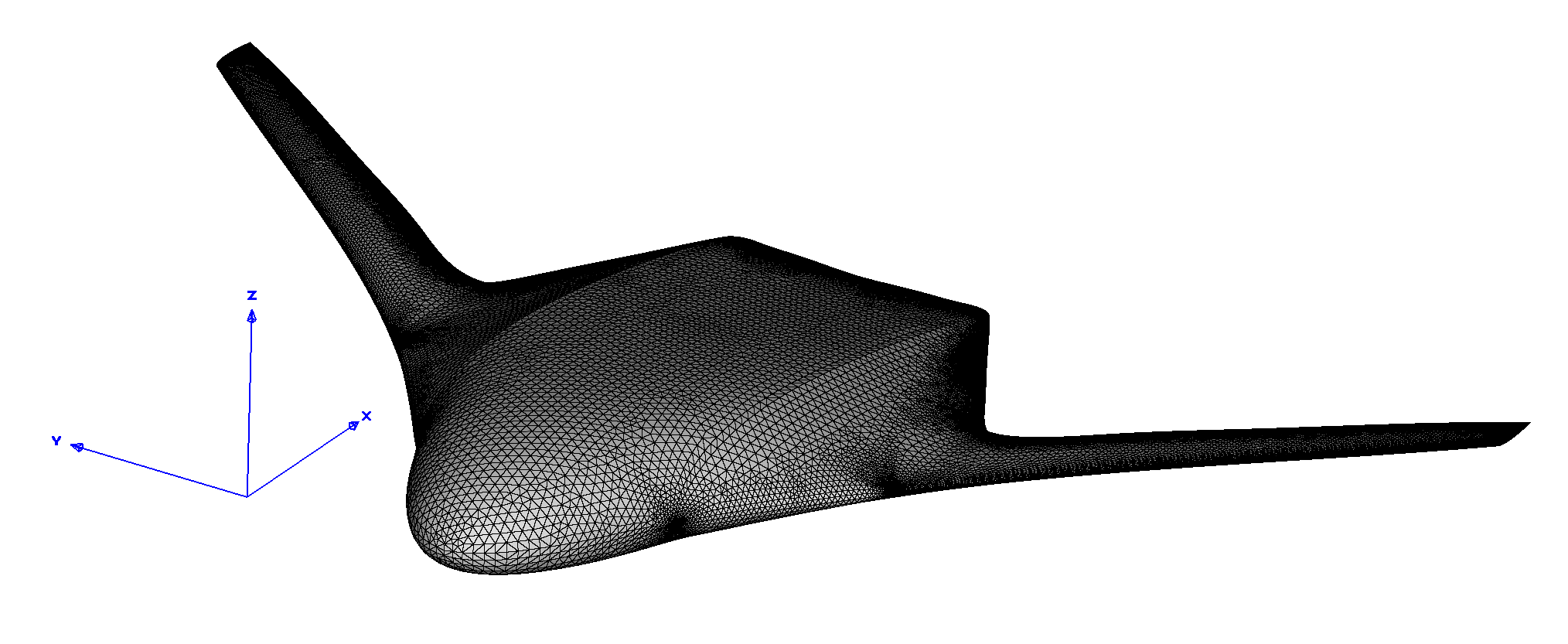} 
        \caption{Surface mesh with coordinate axes.}
        \label{fig:subfigA}
    \end{subfigure}
    \begin{subfigure}{0.4\textwidth} 
        \includegraphics[width=\textwidth]{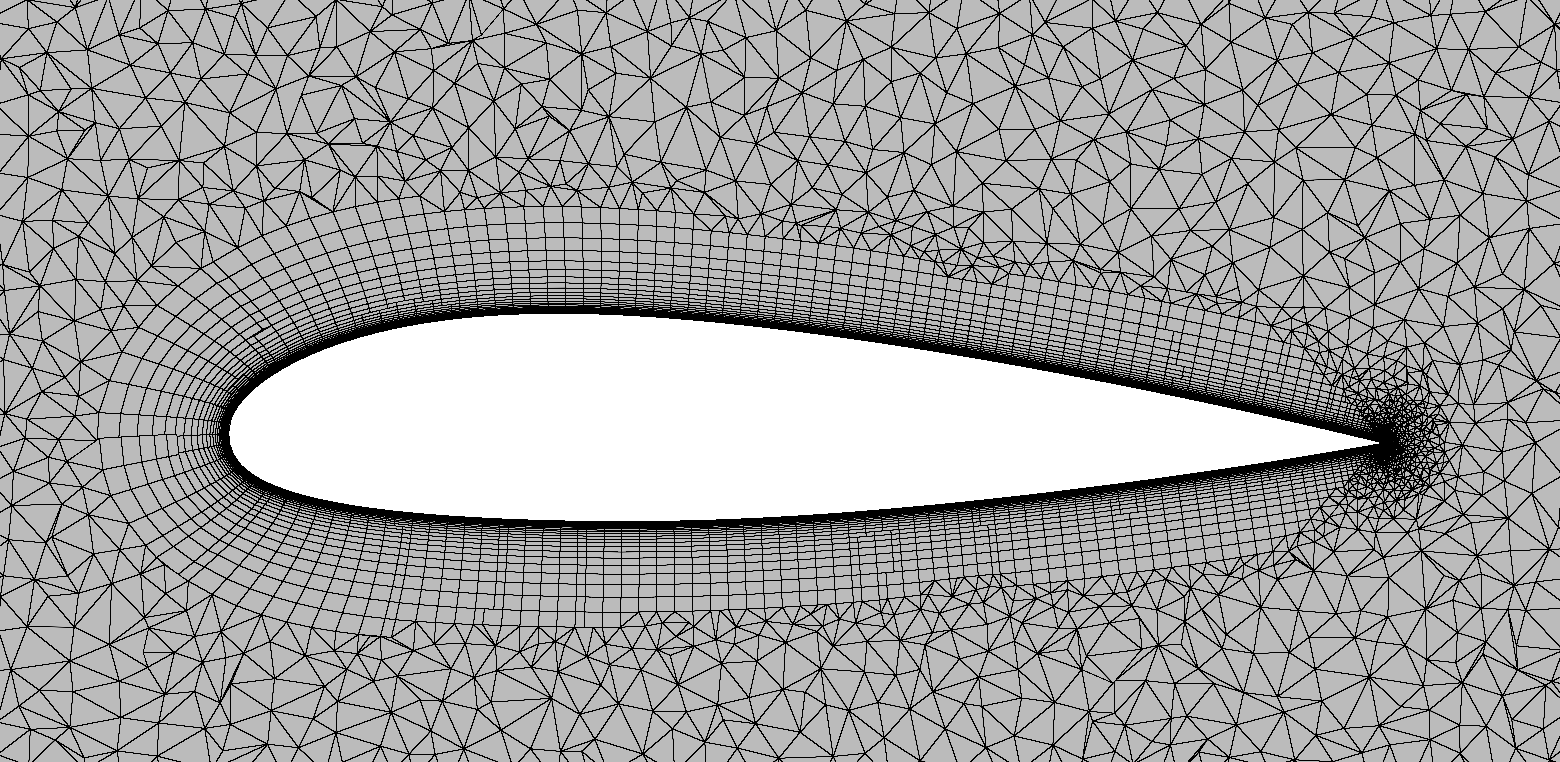} 
        \caption{Boundary layer extrusion along centerline.}
        \label{fig:subfigB}
    \end{subfigure}
    \caption{Example visualization of an automatically generated computational fluid dynamics mesh.}
    \label{fig:meshes}
\end{figure}

CFD meshes were generated using the Pointwise meshing software~\cite{pointwise}.
To facilitate the generation of over 12k meshes, the Pointwise Glyph scripting functionality was leveraged to automate the meshing process.
The surface meshes varied between 65k and 162k cells depending on the specific geometry.
Volume meshes varied between 6.5 and 11.6 million cells and contain mixed element types.
Boundary layer cells were grown off the surface of the aircraft with an initial spacing selected to maintain $y^+$ values less than 1 in order to accurately resolve the boundary layer. 
Example visualizations of a mesh can be seen in Figure~\ref{fig:meshes} along with the coordinate axes used.
The coordinate system origin is located at the nose of the aircraft.

A mesh convergence study was performed by refining the boundary layer and region of the volume near the aircraft.
The selected mesh resolution for this dataset saw error in integrated $C_L$ less than 5\% of the mesh independent value and error in $C_D$ less than 8\%. 
The decision to use this mesh density was made to balance the level of solution accuracy with solver runtime for creating a large dataset, where the total number of cells in the selected mesh is approximately one-eighth of the thoroughly refined mesh leading to approximately 80\% reduction in runtime from the dense mesh.

\subsection*{CFD Simulation}
The aerodynamic data was generated using the NASA FUN3D CFD solver~\cite{anderson2024fun3d}. 
Steady-state simulations were run solving the Reynolds-Averaged Navier-Stokes (RANS) equations using the Spalart-Allmaras turbulence model.
A fully resolved, viscous wall boundary condition was applied on the surface of the aircraft.
A spherical farfield was applied far from the vehicle for inflow/outflow conditions with a turbulence intensity of 3\%.

The parameters to generate the CFD cases were sampled using LHS over altitude, mach (training set), dynamic pressure (test set), vehicle centerline length, and angle of attack.
The ranges of these parameters, as well as their sampling scheme, can be seen in Table~\ref{tab:flightconditions}.

We intentionally simulate each geometry at \emph{one} flight condition (sampled once per geometry) to prioritize \emph{geometric diversity} over repeated condition sweeps. This design targets \emph{geometric generalization} (learning across many distinct planforms) rather than \emph{operational generalization} (off-design behavior for the same geometry). With a fixed compute budget, allocating solves to new geometries yields broader coverage of the design space and supports robust geometry-disjoint evaluation. Future releases may include \emph{multi-condition cohorts} per geometry to study off-design performance and condition coupling.

These sampled parameters allow for the calculation of the simulation flight condition inputs: Mach number, Reynolds number, temperature, and angle of attack. Vehicle length was sampled on a log scale in order to smooth the distribution of the computed Reynolds number.
Reynolds number was not sampled directly, as this approach would lead to too much variation in the effective vehicle centerline length, presenting substantial challenges in automating the mesh generation process, specifically for boundary layer resolution.
All simulations were completed for a model centerline length of 1\,m to simplify the case setup, but the results can be used to predict for the full range of vehicle lengths leveraging the principle of flow similarity (see \cite{andersonaero} for detailed discussion).
Reference area and reference length for force and moment non-dimensionalization were likewise set to 1~m$^2$ and 1~m, respectively, to simplify computations.
The moment reference location is set to the nose of the vehicle, again for simplicity given the large variation of geometries present in the dataset. 

\begin{table}[h]
\centering
\caption{Sampling of flight condition inputs for model training and testing on Group 3 UAS representative conditions.}
\label{tab:flightconditions}
\begin{tabularx}{\linewidth}{@{} l c c c @{}}
\toprule
\textbf{Parameter} & \textbf{Sampling} & \textbf{Training Data} & \textbf{Test Data} \\
\midrule
Altitude [kft] & Linear & [0, 40] & [0, 18] \\
Mach [--] & Linear & [0.05, 0.5] & -- \\
Calibrated Airspeed [kt] & Linear & -- & [33, 250] \\
Centerline Length [m] & Log & [0.1, 10] & [1, 10] \\
Angle of Attack [$^\circ$] & Linear & [-8, 16] & [-8, 16] \\
\bottomrule
\end{tabularx}
\end{table}

The sampled parameters for the training and benchmarking phases are summarized in Table~\ref{tab:flightconditions}. While the training envelope extends to \qty{40000}{ft} to support Group 5 UAS (high-altitude, long-endurance systems), we evaluate our primary benchmarks on a test subset focused on Group 3 UAS. These represent tactical unmanned systems weighing between \num{55} and \qty{1320}{lbs}, operating below \qty{18000}{ft} Mean Sea Level (MSL) at speeds under \qty{250}{knots} \cite{faa_aim_11_3}. Focusing testing on this regime ensures the practical utility of our surrogates for a realistic design task.

\subsection*{Results Processing}
A post-processing routine was used to filter non-converged CFD simulations, collect the results of all the converged cases, and create a VTK file for the distributed surface quantities for each case.
A residual filter of $1\times10^{-8}$ for density and $1\times10^{-6}$ for turbulence was applied, leading to the training and test sets containing 9992 and 2500 cases, respectively.
All converged results were aggregated together to compile the geometry parameters, flight conditions, and integrated lift, drag, and moment coefficients for each run.
These scalar outputs for each case are paired with a VTK file containing the surface mesh with the distribution of pressure coefficient ($C_p$) as well as components of skin friction ($C_{f_x}, C_{f_y}, C_{f_z}$) across the surface of the aircraft.

\section{Dataset Characteristics}
\begin{figure}[h!]
  \centering
  \includegraphics[width=\linewidth]{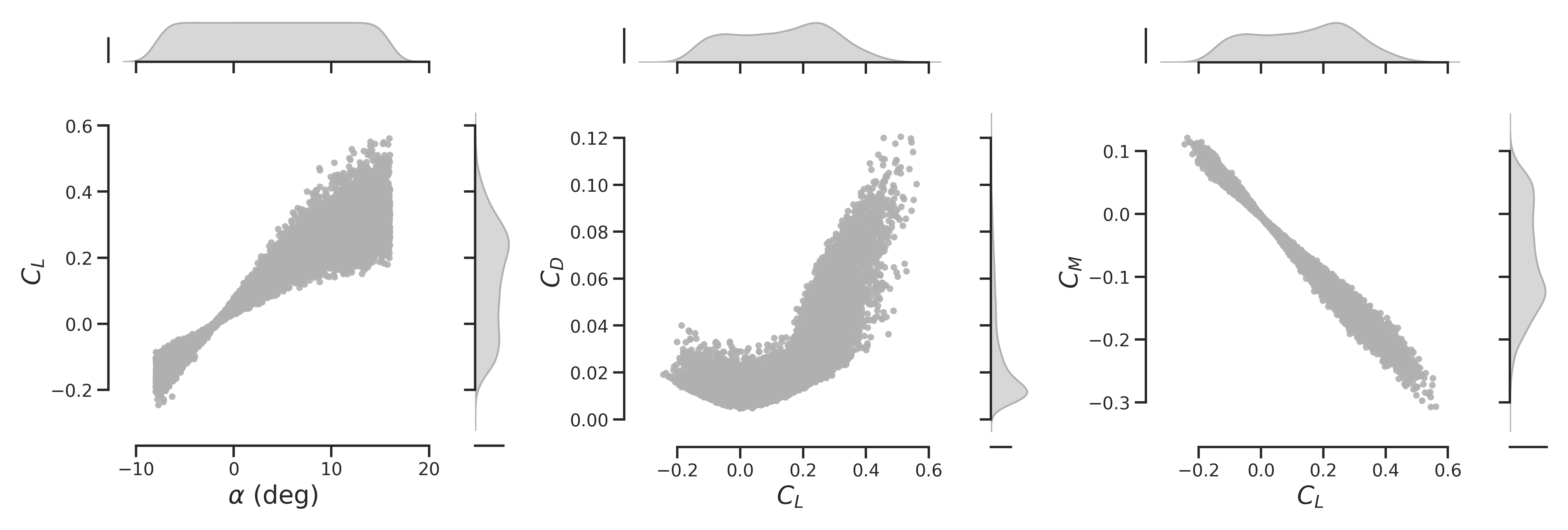}
  \caption{Scatter plots of aerodynamic relationships: (Left) Lift coefficient ($C_L$) vs angle of attack ($\alpha$), (Center) Drag coefficient ($C_D$) vs Lift coefficient ($C_L$), and (Right) Pitching moment coefficient ($C_M$) vs Lift coefficient ($C_L$). These relationships provide insights into aerodynamic performance and longitudinal stability characteristics.}

  \label{fig:coeff_scatter}
\end{figure}

The dataset generation process was performed using in house supercomputing resources.
The breakdown of computational time per task can be found in Table~\ref{tab:datagenerationtime} along with the compute resource utilized for that task. Figure~\ref{fig:airplane_diagonal} illustrates representative geometries from BlendedNet++, showcasing the diversity of fuselage shapes, wing spans, and overall configurations present in the dataset.
The total dataset is roughly 60~GB (uncompressed) where each VTK file is roughly 5~MB.
An example from BlendedNet++ is shown in Figure~\ref{fig:sample_case}, including the baseline 3D geometry and the corresponding surface flow fields (\(C_p\), \(C_{fx}\), and \(C_{fz}\)). To summarize global aerodynamic trends across the dataset, Fig.~\ref{fig:coeff_scatter} visualizes key pairwise relationships.

\begin{table}[h]
\centering
\caption{Computational cost breakdown for the dataset generation process}
\label{tab:datagenerationtime}
\begin{tabular}{ccc}
\toprule
\textbf{Task} & \textbf{Resource} & \textbf{Time per Case} \\
\midrule
Geometry Generation & Xeon-p8 CPU & 10--15 sec \\
Mesh Generation & Xeon-p8 CPU & 3--4 min \\
CFD Solve & 2xV100 GPU & 1.5 hr\\
& 2xH100 GPU & 30--35 min \\
\bottomrule
\end{tabular}
\end{table}

The dataset contains a predefined train/test split of 9992 training cases and 2500 testing cases, representing an 80/20 split.
As previously stated, the geometry parameters and flight conditions were independently sampled via LHS for each piece of the split.

\subsection*{Comparison to BlendedNet}
The BlendedNet++ dataset introduced in this work is an improvement on the previously released BlendedNet dataset~\cite{blendednet}.
Several enhancements to the dataset generation process distinguish this data from the previous work.
The automated meshing routine was improved, leading to better quality meshes and improved convergence of the CFD simulations. 
The $X_3$ parameter has been introduced to provide better control over the location of the wing root in the chord-wise direction (as opposed to using a sweep angle for the leading edge of that section).
The simulated angle of attack range was narrowed from $[-10,20]^\circ$ to $[-8,16]^\circ$ to increase the density of sampling for more realistic aircraft flight conditions.
While BlendedNet++ contains only roughly 2.5k more CFD cases than BlendedNet, each CFD case is for a unique geometry, increasing the diversity of the geometry parameter space. Table~\ref{tab:blendednet_comparison} highlights some of these changes and contains the specific geometry and CFD case counts between the datasets.

\begin{table}[h]
\centering
\caption{Comparison between BlendedNet and BlendedNet++ datasets}
\label{tab:blendednet_comparison}
\begin{tabular}{ccc}
\toprule
\textbf{Quantity} & \textbf{BlendedNet} & \textbf{BlendedNet++} \\
\midrule
Total Geometries & 1,099 & 12,492 \\
Training CFD Cases & 8,830 & 9,992 \\
Testing CFD Cases & 870 & 2,500 \\
Flight Condition per Geometry & 8--9 & 1 \\
\bottomrule
\end{tabular}
\end{table}


\section{Surrogate Model Development}

\paragraph{Why a five-model benchmark?}
It is often unclear which representation works best for \emph{field-level} surface prediction in aerodynamics, and reported ``best models'' vary significantly by dataset, resolution, and metric~\cite{dong2021deep,sabater2022fast,elrefaie2024drivaernet++, elrefaie2025drivaernet}. Recent benchmark efforts and editorials argue that progress requires common datasets, geometry-disjoint splits, and unified metrics to separate modeling choices from data curation~\cite{moseley2021airfrans, jmd_datasets_editorial_2024, elrefaie2025carbench}. We therefore evaluate five complementary families that cover distinct inductive biases while keeping preprocessing minimal and comparable: point-set, hierarchical graph encoders, operator-style transformers, coordinate transformers, and FiLM-style hypernetwork conditioning. To isolate architectural effects, we hold inputs, losses, and evaluation protocols fixed, and implement a shared training loop via the \emph{NeuralSolver} codebase~\cite{neuralsolver_library}. All five forward surrogates are trained and evaluated on a single NVIDIA A100 GPU with a shared early-stopping schedule and a maximum wall-clock budget of roughly 48 hours per model.

We benchmark these five forward models for \emph{pointwise} prediction of $C_p$ and $\mathbf{C}_f$: \textit{PointNet}, \textit{Transolver}, \textit{GraphUNet}, \textit{GNOT}, and \textit{FiLMNet}. Except for FiLMNet, all models are implemented and trained using \emph{NeuralSolver}. FiLMNet follows the BlendedNet architecture~\cite{blendednet}. We omit the spanwise skin-friction $C_{f_y}$ because the zero-sideslip, symmetric BWB setup implies near-zero integrated side force, making $C_{f_y}$ negligible for $C_L$ and $C_D$ relative to $C_p$, $C_{f_x}$, and $C_{f_z}$. The dataset still includes $C_{f_y}$ for completeness. We report exact train/test counts and construct geometry-disjoint splits using independent LHS samples for each split. All normalization statistics (e.g., mean/std standardization) are computed on the training split only and applied to the test split.

\begin{figure}[htbp]
    \centering
    \includegraphics[width=0.8\textwidth]{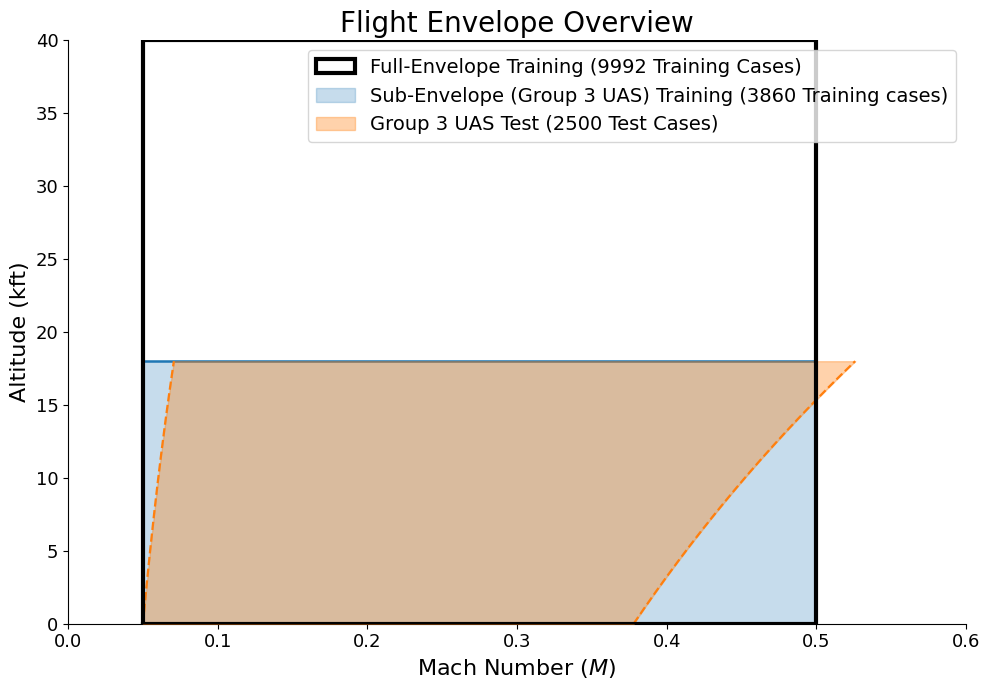}
    \caption{Flight envelope overview showing the Full-Envelope training bounds (40~kft), the Sub-Envelope training region ($\le$18~kft), and the specific Group 3 UAS test cases.}
    \label{fig:flight_envelope}
\end{figure}

\subsection{Training Regimes and Data Selection}
To investigate the trade-off between broad physical coverage and specialized operational accuracy, we establish two distinct training regimes:
\begin{enumerate}
    \item \textbf{Full-Envelope Regime:} Models are trained on the complete training split of \num{9992} geometries spanning the full altitude range up to \qty{40}{kft}. This regime represents a general approach intended to capture a wide range of UAS operations (Groups 1--5).
    \item \textbf{Sub-Envelope (Group 3 UAS) Regime:} Models are trained on a filtered subset of the full- envelope training data containing only cases where the altitude is $\leq \qty{18}{kft}$, resulting in a specialized training set of \num{3860} cases. 
\end{enumerate}

Both regimes are evaluated on a shared \textbf{Group 3 UAS Test Set} ($N=\num{2500}$) where all flight conditions are sampled within the tactical envelope shown in Figure~\ref{fig:flight_envelope} and Table~\ref{tab:flightconditions}. This allows us to see if including high-altitude data actually makes the model less accurate for tactical missions—a case where more data might lead to lower precision for specific tasks.

\subsection{Feature-wise Linear Modulation Network (FiLMNet)}
The FiLMNet predicts
\[
\mathbf{u}_i
=\bigl[C_p(\mathbf{x}_i,\mathbf{n}_i),\;C_{f_x}(\mathbf{x}_i,\mathbf{n}_i),\;C_{f_z}(\mathbf{x}_i,\mathbf{n}_i)\bigr]\in\mathbb{R}^3
\]
at each surface point $i$ from a base MLP that takes \emph{only} the pointwise geometry
\[
\mathbf{s}_i \;=\; [\,\mathbf{x}_i,\;\mathbf{n}_i\,]\in\mathbb{R}^{6},
\]
while the layer-wise modulation is conditioned on the 12 broadcast variables
\[
\boldsymbol{\mu}=\big[\,Re_L,\,M_\infty,\,\alpha,\,\mathrm{B1},\mathrm{B2},\mathrm{B3},\mathrm{C2},\mathrm{C3},\mathrm{C4},\mathrm{S1},\mathrm{S3},\mathrm{X3}\bigr]\in\mathbb{R}^{12}.
\]
Let $h_0=\mathbf{s}_i$ and $h_\ell$ be the hidden activation at layer $\ell$ of the base MLP. A small hypernetwork $h_\psi$ maps $\boldsymbol{\mu}$ to FiLM parameters $(\boldsymbol{\gamma}_\ell,\boldsymbol{\beta}_\ell)$ per layer, and we apply FiLM as
\[
\tilde{h}_\ell \;=\; \boldsymbol{\gamma}_\ell(\boldsymbol{\mu}) \odot h_\ell \;+\; \boldsymbol{\beta}_\ell(\boldsymbol{\mu}),\qquad
h_{\ell+1} \;=\; \sigma\!\bigl(W_\ell \tilde{h}_\ell + b_\ell\bigr),
\]
with $\odot$ denoting elementwise multiplication and $\sigma$ a pointwise nonlinearity. The output head maps the final activation to $\mathbf{u}_i\in\mathbb{R}^3$.

\subsection{PointNet}
PointNet \cite{qi2017pointnet} is a deep-learning architecture that directly processes 3D point clouds by applying shared multilayer perceptrons (MLPs) to each point and then aggregating features via a permutation-invariant operator (e.g., max pooling).

\begin{equation}
\label{eq:fx-def}
\mathrm{fx}=
\left[
\begin{aligned}
&\underbrace{Re_L,\; M_\infty,\; \alpha}_{\text{3 flight conditions}},\\
&\underbrace{B_1/C_1,\; B_2/C_1,\; B_3/C_1,\; C_2/C_1,\; C_3/C_1,\; C_4/C_1,\; S_1,\; S_3,\; X_3/C_1}_{\text{9 geometric parameters}},\\
&\underbrace{n_x,\;n_y,\;n_z}_{\text{surface normals}}
\end{aligned}
\right].
\end{equation}

\textbf{Our adaptation.} Each point supplies its 3D coordinates $(x,y,z)$ and a 15-D feature vector: the 9 planform parameters and 3 flight conditions (broadcast to all points) plus the 3 local surface normals $(n_x,n_y,n_z)$. We concatenate coordinates and features at the input and regress $(C_p,\mathbf{C}_f)$.

\subsection{Transolver}
Transolver introduces Physics-Attention to improve neural operators on irregular meshes by aggregating points into physics-aware tokens and attending over tokens. 
It has been reported to perform competitively on irregular-mesh partial differential equation (PDE) surrogate tasks and flow problems, with strong accuracy for long-range interactions and near-linear scaling in the number of surface points due to efficient attention approximations~\cite{wu2024Transolver}.
It captures nonlocal coupling via attention, works directly on point clouds without remeshing or surface-coordinate parameterization, scales to high-resolution surfaces through tokenization, and ships with a maintained open-source implementation that supports reproducibility~\cite{wu2024Transolver}. We treat Transolver as a representative of coordinate and point–token transformer-based models for field prediction.

\textbf{Our adaptation.} The current Transolver implementation targets fixed PDE setups and does not adjust for varying boundary conditions or changing geometries as inputs. Our prediction task varies \emph{both}, so we condition the model explicitly by concatenating the flight conditions and planform geometries as additional features to every surface node. We use the same input preparation as in Eq.~\eqref{eq:fx-def}: coordinates $\mathbf{x}\in\mathbb{R}^{N\times 3}$ and per-point features $\mathrm{fx}\in\mathbb{R}^{N\times 15}$ (9 planform parameters and 3 flight conditions plus the 3 local surface normals $(n_x,n_y,n_z)$), concatenated at the input. The concatenated node features are then processed by Physics-Attention so flight conditions and geometric features inform the slice tokens.



\subsection{GraphUNet}
GraphUNet~\cite{gao2019graphunet} extends the U-Net architecture to graphs by using the connectivity of the original CFD surface mesh through graph pooling and unpooling layers. This structure allows the network to capture both local and global dependencies while preserving resolution for pointwise predictions.

\textbf{Our adaptation.} We follow the same input preparation as in Eq.~\eqref{eq:fx-def}, providing coordinates $\mathbf{x}\in\mathbb{R}^{N\times 3}$ and the per-point feature vector $\mathrm{fx}\in\mathbb{R}^{N\times 15}$, concatenated at the input. By embedding the 9 planform parameters and 3 flight conditions alongside the 3 local surface normals into every node, the hierarchical encoder can reason about global planform trends and local pressure gradients simultaneously. The model is trained to predict $(C_p, C_f)$ fields, utilizing the multi-scale inductive bias of the U-Net structure to resolve complex boundary layer interactions across the BWB airframe.

\subsection{GNOT (General Neural Operator Transformer)}
GNOT ~\cite{hao2023gnot} is a transformer-based framework that uses linear attention and geometric gating to efficiently learn PDE operators across irregular meshes, multiple inputs, and multi-scale systems.

\textbf{Our adaptation.} We follow the same input preparation as in Eq.~\eqref{eq:fx-def}, providing coordinates $\mathbf{x}\in\mathbb{R}^{N\times 3}$ and the per-point feature vector $\mathrm{fx}\in\mathbb{R}^{N\times 15}$, concatenated at the input. GNOT takes $(\mathbf{x},\mathrm{fx})$ as input to predict $(C_p,C_f)$, using global attention to capture nonlocal aerodynamic coupling.

\subsection{Training Objective}
We predict three per-point channels: (i) pressure coefficient $C_p$, (ii) skin-friction in the streamwise direction $C_{f_x}$, and (iii) skin-friction in the vertical direction $C_{f_z}$. We train on their normalized forms $\tilde C=(C-\mu)/\sigma$. For $N$ surface points, the loss is
\begin{equation}
\mathcal{L}
=\frac{1}{N}\sum_{i=1}^{N}\Big[
\big(\hat{\tilde C}_{p,i}-\tilde C_{p,i}\big)^2
+\big(\hat{\tilde C}_{f_x,i}-\tilde C_{f_x,i}\big)^2
+\big(\hat{\tilde C}_{f_z,i}-\tilde C_{f_z,i}\big)^2
\Big],
\end{equation}
where hats denote predictions and tildes denote normalized targets.

This multi-field learning formulation which predicts all fields within a single unified network is chosen to exploit the inherent physical coupling between pressure and skin friction. In fluid mechanics, surface pressure gradients $\nabla C_p$ directly influence boundary layer development and the resulting skin-friction distributions. By utilizing a shared latent representation, we hypothesize that the models are able to learn a physically consistent internal mapping of flow phenomena, such as stagnation points and separation bubbles, which characterize both $C_p$ and $C_f$ fields. Architecturally, this joint prediction acts as a form of inductive regularization, preventing the model from overfitting to a single coefficient at the expense of global physical validity. Furthermore, a single-model approach significantly enhances computational efficiency during both training and inverse-design optimization, providing a unified gradient source for the downstream generative pipeline.

\subsection{Hyperparameters}
All models (except FiLMNet) were trained for a maximum of 1,000 epochs with an early stopping patience of 100 epochs. FiLMNet was trained for up to 10,000 epochs with a 1000 epoch patience. Table~\ref{tab:model_configs} summarizes the core hyperparameters used for the benchmark.

\begin{table}[htbp]
\centering
\caption{Training and model configurations for the BlendedNet++ benchmark architectures. All models are trained using Mean Squared Error (MSE) loss on surface point clouds. The \textit{Embed} column represents the hidden layer width or embedding dimension. \textit{Concat} indicates that features are concatenated at every point, while \textit{Condition} refers to FiLM-based hypernetwork modulation.}
\label{tab:model_configs}
\begin{adjustbox}{width=\textwidth,center}
\begin{tabular}{llcccccccl}
\toprule
\textbf{Model} & \textbf{Conditioning} & \textbf{Max Epochs} & \textbf{Patience} & \textbf{LR} & \textbf{Opt} & \textbf{Batch} & \textbf{Layers} & \textbf{Heads} & \textbf{Embed} \\
\midrule
FiLMNet & Flight + Geom (FiLM) & \num{10000} & 1000 & \num{5e-4} & Adam & 64 & 4 & -- & 256 \\
PointNet & Flight + Geom (Concat) & \num{1000} & 100 & \num{1e-3} & Adam & 64 & 8 & 8 & 32 \\
Transolver & Flight + Geom (Concat) & \num{1000} & 100 & \num{1e-3} & Adam & 64 & 8 & 8 & 256 \\
GNOT & Flight + Geom (Concat) & \num{1000} & 100 & \num{1e-3} & Adam & 64 & 8 & 8 & 128 \\
GraphUNet & Flight + Geom (Concat) & \num{1000} & 100 & \num{1e-3} & Adam & 32 & 4 & -- & 128 \\
\bottomrule
\end{tabular}
\end{adjustbox}
\end{table}

For test metrics across all model, we again compare the checkpoint with the best validation loss and the final epoch checkpoint and we report the better result.

\section{Inverse Design via Conditional Diffusion}
\label{sec:inverse}
We pose inverse design as conditional generation \(p(\mathbf{p}\mid \boldsymbol{\mu})\) where \(\mathbf{p}\in\mathbb{R}^{9}\) lies in a box \(\mathcal{B}\) (Table~\ref{tab:param_bounds}). The conditioning vector is defined as:
\[
\boldsymbol{\mu}=\big[\text{alt}_{\text{ft}}, \log_{10}(\text{Re}_L), M_\infty, \alpha, (L/D)_{\text{target}}\big]\in\mathbb{R}^{5}.
\]
While \(\text{Re}_L\) is physically dependent on altitude and Mach number, both are included in \(\boldsymbol{\mu}\) to provide the model with an explicit scaling parameter for viscous effects, which was found to improve training convergence. A denoising diffusion model parameterizes the learned reverse distribution \(p_\theta(\mathbf{x}_{t-1} \mid \mathbf{x}_t, \boldsymbol{\mu})\), which approximates the true posterior \(q(\mathbf{x}_{t-1} \mid \mathbf{x}_t)\). Samples are projected onto \(\mathcal{B}\) following the generation process. Full architectural choices and training details are provided in the Appendix.

\subsection{Model}
We employ a residual MLP denoiser \(f_\theta\) that predicts additive noise \(\hat{\boldsymbol{\epsilon}}\) given a noised geometry vector \(\mathbf{x}_t\), diffusion timestep \(t\), and condition \(\boldsymbol{\mu}\):
\[
\hat{\boldsymbol{\epsilon}}=f_\theta(\mathbf{x}_t,t,\boldsymbol{\mu}).
\]
The timestep \(t\) and \(\boldsymbol{\mu}\) are embedded by separate linear projections and concatenated with \(\mathbf{x}_t\). To avoid notation ambiguity with the angle of attack (\(\alpha\)), we denote the diffusion coefficients as \(a\). A cosine \(\beta_t\) schedule defines \(a_t=1-\beta_t\) and the cumulative product \(\bar{a}_t=\prod_{s=1}^t a_s\).

\subsection{Forward (Noising) Process}
The training target \(\mathbf{x}_0\) is the normalized geometry vector. We draw noise \(\boldsymbol{\epsilon}\sim\mathcal{N}(\mathbf{0},\mathbf{I})\) and sample the forward process as:
\[
q(\mathbf{x}_t\mid \mathbf{x}_0)=\mathcal{N}\!\left(\sqrt{\bar{a}_t}\,\mathbf{x}_0,\; (1-\bar{a}_t)\mathbf{I}\right),
\quad
\mathbf{x}_t=\sqrt{\bar{a}_t}\,\mathbf{x}_0+\sqrt{1-\bar{a}_t}\,\boldsymbol{\epsilon}.
\]

\subsection{Training Objective}
We optimize the standard noise-prediction loss
\[
\mathcal{L}(\theta)=\mathbb{E}_{\mathbf{x}_0,\boldsymbol{\mu},t,\boldsymbol{\epsilon}}
\!\left[\left\|\boldsymbol{\epsilon}-f_\theta\!\big(\mathbf{x}_t,t,\boldsymbol{\mu}\big)\right\|_2^2\right],
\]
which is equivalent (up to constants) to a variational bound when the model predicts \(\boldsymbol{\epsilon}\). Cosine \(\beta_t\), sinusoidal \(t\)-embeddings, and residual MLP blocks follow common practice for conditional denoising diffusion probabilistic models (DDPMs).

\subsection{Conditioning and Normalization}
We standardize \(\boldsymbol{\mu}\) (with \(\log_{10}(\text{Re}_L)\) as an input) using a fitted $\mathcal{N}(0,1)$ scaler. Geometry parameters are scaled to \([-1,1]\) during training and denormalized only at sampling.

\subsection{Sampling}
Given a condition \(\boldsymbol{\mu}\), we first standardize it, then draw \(\mathbf{x}_T\sim\mathcal{N}(\mathbf{0},\mathbf{I})\) and iterate for \(t=T,\ldots,1\):
\[
\hat{\mathbf{x}}_0
=\frac{1}{\sqrt{\bar{a}_t}}\Big(\mathbf{x}_t - \sqrt{1-\bar{a}_t}\,\hat{\boldsymbol{\epsilon}}_\theta(\mathbf{x}_t,t,\boldsymbol{\mu})\Big).
\]
The reverse mean and variance are:
\begin{equation}
\begin{aligned}
\boldsymbol{\mu}_\theta(\mathbf{x}_t,t,\boldsymbol{\mu})
&=\frac{\beta_t \sqrt{\bar{a}_{t-1}}}{1-\bar{a}_t}\,\hat{\mathbf{x}}_0
+\frac{(1-\bar{a}_{t-1})\sqrt{a_t}}{1-\bar{a}_t}\,\mathbf{x}_t,\\[2pt]
\sigma_t^2
&=\frac{\beta_t(1-\bar{a}_{t-1})}{1-\bar{a}_t}.
\end{aligned}
\end{equation}

We then sample
\[
\mathbf{x}_{t-1}=
\begin{cases}
\boldsymbol{\mu}_\theta(\mathbf{x}_t,t,\boldsymbol{\mu})+\sigma_t\,\mathbf{z}, & t>1,\ \mathbf{z}\sim\mathcal{N}(\mathbf{0},\mathbf{I}),\\[2pt]
\boldsymbol{\mu}_\theta(\mathbf{x}_t,t,\boldsymbol{\mu}), & t=1.
\end{cases}
\]
Finally, we invert the geometry scaler to obtain physical units \(\mathbf{p}=\mathrm{denorm}(\mathbf{x}_0)\). For each \(\boldsymbol{\mu}\), we draw \(K=100\) i.i.d.\ samples to quantify diversity and performance (Section~\ref{sec:results-inverse}).

\section{Results and Discussion}

\subsection{Forward Benchmark Results}

In this section, we present the pointwise field prediction errors for the primary aerodynamic surface coefficients ($C_p, C_{f_x}, C_{f_z}$). It is critical to note that $C_p$ values are typically two orders of magnitude larger than the skin friction components, making the relative error in pressure the most significant metric for overall aerodynamic performance (lift and drag).

\begin{table}[htbp]
\centering
\caption{Forward prediction errors on the \emph{unseen test} set: Group 3 UAS Flight Regime (alt $<$ \qty{18}{kft}). All models are trained on the \textbf{Full-Envelope} (complete 40k ft dataset with 9992 training cases)}

\label{tab:full_envelope_errors}
\scriptsize
\setlength{\tabcolsep}{2.5pt}
\renewcommand{\arraystretch}{1.1}

\begin{adjustbox}{width=\textwidth}
\begin{tabular}{l c *{4}{S[table-format=2.2]} *{4}{S[table-format=2.2]} *{4}{S[table-format=2.2]}}
\toprule
& & \multicolumn{4}{c}{$C_p$} & \multicolumn{4}{c}{$C_{f_x}$} & \multicolumn{4}{c}{$C_{f_z}$} \\
\cmidrule(lr){3-6}\cmidrule(lr){7-10}\cmidrule(l){11-14}
\textbf{Model} & \textbf{Params}
& {MSE $(\times 10^{-2})$} & {MAE $(\times 10^{-2})$} & {RelL1} & {RelL2}
& {MSE $(\times 10^{-5})$} & {MAE $(\times 10^{-3})$} & {RelL1} & {RelL2}
& {MSE $(\times 10^{-5})$} & {MAE $(\times 10^{-4})$} & {RelL1} & {RelL2} \\
\midrule
FiLMNet    & 0.80M & \textbf{0.36} & \textbf{1.99} & \textbf{0.06} & \textbf{0.10} & \textbf{0.45} & \textbf{0.56} & \textbf{0.13} & \textbf{0.32} & \textbf{0.12} & \textbf{3.11} & \textbf{0.18} & \textbf{0.21} \\
PointNet   & 0.92M & 9.09 & 18.89 & 0.61 & 0.53 & 3.92 & 4.53 & 1.08 & 0.95 & 1.17 & 16.16 & 0.95 & 0.66 \\
Transolver & 2.82M & 2.25 & 9.52 & 0.31 & 0.26 & 1.82 & 3.66 & 0.87 & 0.65 & 0.29 & 5.37 & 0.31 & 0.33 \\
GNOT       & 2.90M & 14.22 & 20.99 & 0.68 & 0.66 & 2.62 & 3.95 & 0.94 & 0.78 & 1.32 & 13.06 & 0.77 & 0.70 \\
GraphUNet  & 56.67M & 1.14 & 5.10 & 0.16 & 0.19 
& 0.76 & 0.92 & 0.22 & 0.42 
& 0.28 & 5.80 & 0.34 & 0.32 \\
\bottomrule
\end{tabular}
\end{adjustbox}
\end{table}

\begin{table}[htbp]
\centering
\caption{Forward prediction errors on the \emph{unseen test} set: Group 3 UAS Flight Regime (alt $<$ \qty{18}{kft}). All models are trained specifically on the \textbf{Sub-Envelope} (Group 3 UAS regime < 18k ft, with 3860 training cases)}
\label{tab:feasible_envelope_errors}
\scriptsize
\setlength{\tabcolsep}{2.5pt}
\renewcommand{\arraystretch}{1.1}

\begin{adjustbox}{width=\textwidth}
\begin{tabular}{l c *{4}{S[table-format=2.2]} *{4}{S[table-format=2.2]} *{4}{S[table-format=2.2]}}
\toprule
& & \multicolumn{4}{c}{$C_p$} & \multicolumn{4}{c}{$C_{f_x}$} & \multicolumn{4}{c}{$C_{f_z}$} \\
\cmidrule(lr){3-6}\cmidrule(lr){7-10}\cmidrule(l){11-14}
\textbf{Model} & \textbf{Params}
& {MSE $(\times 10^{-2})$} & {MAE $(\times 10^{-2})$} & {RelL1} & {RelL2}
& {MSE $(\times 10^{-5})$} & {MAE $(\times 10^{-3})$} & {RelL1} & {RelL2}
& {MSE $(\times 10^{-5})$} & {MAE $(\times 10^{-4})$} & {RelL1} & {RelL2} \\
\midrule
FiLMNet    & 0.80M & 1.42 & 6.52 & 0.15 & 0.15 & 0.76 & 2.48 & 0.59 & 0.42 & 0.08 & 3.25 & 0.15 & 0.13 \\
PointNet   & 0.92M & 0.35 & 2.93 & 0.07 & 0.07 & 0.12 & 0.39 & 0.09 & 0.17 & 0.04 & 2.53 & 0.11 & 0.09 \\
Transolver & 2.82M & \textbf{0.27} & \textbf{1.75} & \textbf{0.04} & \textbf{0.06} & \textbf{0.12} & \textbf{0.32} & \textbf{0.08} & \textbf{0.17} & \textbf{0.03} & \textbf{1.88} & \textbf{0.08} & \textbf{0.08} \\
GNOT       & 2.90M & 6.44 & 11.73 & 0.27 & 0.31 & 0.42 & 1.03 & 0.25 & 0.32 & 0.24 & 6.41 & 0.28 & 0.22 \\
GraphUNet  & 56.67M & 1.08 & 4.63 & 0.15 & 0.18 
& 0.41 & 0.70 & 0.17 & 0.31 
& 0.26 & 4.88 & 0.28 & 0.31 \\
\bottomrule
\end{tabular}
\end{adjustbox}
\end{table}

\subsubsection{Specialization vs. Generalization: The Training Regime Paradox}
A core finding of this benchmark is that \textbf{more data does not necessarily yield better performance.} As shown in Tables~\ref{tab:full_envelope_errors} and \ref{tab:feasible_envelope_errors}, all models except FiLMNet trained on the sub-envelope (Group 3 UAS) regime consistently outperformed their Full-Envelope counterparts when evaluated on the unseen test set, despite having access to only \num{3860} training cases compared to \num{9992} in the Full-Envelope.

The most dramatic leap was observed in the Transolver architecture, where pressure MAE improved nearly five-fold (9.52 to 1.75) when the training distribution was refined. This suggests that for complex BWB geometries, localized "physics-attention" mechanisms are highly sensitive to the range of flow conditions. We hypothesize that when tasked with a narrower envelope, these models can dedicate their capacity to the specific viscous interactions and pressure gradients unique to low altitude flight. 

Conversely, FiLMNet exhibits remarkable stability across regimes. We hypothesize that these models leverage specific architectural inductive biases that anchor predictions in global features rather than local mapping shortcuts. FiLMNet effectively factorizes physics from geometry via its hypernetwork, which treats flight conditions as a global operator that scales and shifts layer in the base MLP.

\subsubsection{Physics of Error: Pressure Dominance and Viscous Challenges} There is a clear divide in how models handle different physical field types. Surface pressure ($C_p$) exhibits high global correlation and is well-resolved by most specialized architectures. However, skin friction ($C_f$) presents a steeper challenge for relative accuracy. High relative errors in the Full-Envelope models indicate that viscous effects are often treated as secondary "noise" by the optimizer when the loss signal is dominated by the higher-magnitude pressure field. Only in the Sub-Envelope regime do models like PointNet and Transolver resolve the skin friction with high precision ($RelL1 < 0.10$). This indicates that capturing the subtle boundary-layer behavior of a BWB, especially the spanwise flow components, requires a model that is not "distracted" by the vastly different pressure physics found in high-altitude, high-speed regimes.

\subsubsection{Inductive Bias: FiLM vs. Attention}
The results highlight a fundamental difference in inductive biases:
\begin{itemize}
\item \textbf{Global Feature Modulation (FiLMNet):} FiLM-based hypernetworks prove remarkably robust across the full flight envelope. While they lose their "lead" in the Sub-Envelope regime, they maintain the lowest overall error in the Full-Envelope case. This suggests that feature-wise modulation is the most effective mechanism for adapting a representation to different operating conditions.
\item \textbf{Contextual Physics-Attention (Transolver):} By tokenizing geometry and physics features, Transolver is best for precision in the Group 3 regime. Its ability to capture nonlocal aerodynamic coupling makes it superior for resolving stagnation zones and trailing-edge flow, provided it is trained on a distribution that matches the test set.
\end{itemize}

\subsection{Computational Efficiency and Inference Statistics}

Inference efficiency is also a critical factor to consider. Table~\ref{tab:inference_stats} summarizes the computational footprint of the evaluated architectures. We report Peak Memory Allocated, total FLOPs, and the point-wise throughput (points per second).

\begin{table}[H]
\centering
\caption{Inference computational statistics for candidate architectures.}
\label{tab:inference_stats}
\scriptsize
\begin{tabular}{l c c c S[table-format=9.0]}
\toprule
\textbf{Model} & \textbf{Params (M)} & \textbf{Memory (MB)} & \textbf{GFLOPs} & \textbf{Throughput (pts/sec)} \\
\midrule
FiLMNet    & 0.80 & 76.54  & 0.79   & 4362188 \\
PointNet   & 0.92 & 110.12 & 1.54   & 5218941 \\
Transolver & 2.82 & 116.36 & 15.02  & 2366633 \\
GNOT       & 2.90 & 148.91 & 18.22  & 1542118 \\
GraphUNet  & 56.67 & 438.74 & 232.13  & 145074 \\
\bottomrule
\end{tabular}
\end{table}

While Transolver provides the highest accuracy, it incurs a significant computational penalty. Its GFLOPs (15.02) are nearly 20 times higher than those of FiLMNet. For onboard integration, \textbf{FiLMNet} offers the optimal "sweet spot"—it is the most memory-efficient (76.54 MB) while maintaining competitive accuracy across all regimes. Conversely, if high throughput is required for large-scale design sweeps, \textbf{PointNet}’s simple MLP-based structure delivers the highest processing speed ($>5 \times 10^6$ pts/sec).

\subsection{Statistical Error Distributions and Failure-Case Analysis}

Systematically analyzing the worst-performing cases for each model is crucial to identify where the predictions break down and which flow regions, such as high-curvature wing-body blends, remain most challenging. To provide a detailed assessment beyond average errors, we report percentile-based error statistics ($P_{50}$, $P_{90}$, $P_{95}$, and $P_{99}$) which capture accuracy across typical and extreme flow regions. 

The $P_{50}$ error represents the median deviation per surface point, reflecting typical predictive fidelity in low-gradient regions. The $P_{90}$ and $P_{95}$ percentiles capture the upper tail, corresponding to complex aerodynamic regions such as leading edges and stagnation zones. Finally, the $P_{99}$ error identifies rare but large outliers, typically occurring near sharp discontinuities or the trailing edge of the BWB.

\begin{table}[htbp]
\centering
\caption{Statistical error distribution of surface pressure ($C_p$) and skin friction ($C_{f_x}, C_{f_z}$) prediction across models. Percentile errors are reported in absolute units ($\times 10^{-2}$ for $C_p$, $\times 10^{-4}$ for $C_{f_x}$, and $\times 10^{-4}$ for $C_{f_z}$).}
\label{tab:percentile_errors}
\scriptsize
\begin{adjustbox}{width=\textwidth}
\begin{tabular}{l l S S S S | S S S S | S S S S}
\toprule
& & \multicolumn{4}{c}{$C_p$ Error ($\times 10^{-2}$)} & \multicolumn{4}{c}{$C_{f_x}$ Error ($\times 10^{-4}$)} & \multicolumn{4}{c}{$C_{f_z}$ Error ($\times 10^{-4}$)} \\
\cmidrule(lr){3-6}\cmidrule(lr){7-10}\cmidrule(lr){11-14}
\textbf{Model} & \textbf{Training Set} & {$P_{50}$} & {$P_{90}$} & {$P_{95}$} & {$P_{99}$} & {$P_{50}$} & {$P_{90}$} & {$P_{95}$} & {$P_{99}$} & {$P_{50}$} & {$P_{90}$} & {$P_{95}$} & {$P_{99}$} \\
\midrule
\textbf{FiLMNet}    & Full-Env & \textbf{0.83} & \textbf{3.74} & \textbf{6.51} & \textbf{21.86} & \textbf{1.57} & \textbf{10.36} & \textbf{20.66} & \textbf{74.63} & \textbf{0.43} & \textbf{6.56} & \textbf{15.11} & \textbf{49.51} \\
PointNet   & Full-Env & 10.51 & 46.77 & 65.20 & 109.63 & 35.39 & 76.06 & 111.01 & 229.80 & 4.52 & 43.82 & 67.50 & 145.93 \\
Transolver & Full-Env & 6.44 & 20.25 & 28.06 & 59.70 & 33.28 & 42.21 & 52.85 & 111.11 & 1.27 & 11.05 & 24.35 & 79.15 \\
GNOT       & Full-Env & 11.56 & 46.05 & 72.34 & 163.87 & 33.28 & 61.55 & 86.25 & 176.05 & 3.20 & 30.55 & 57.89 & 168.53 \\
GraphUNet  & Full-Env & 2.42 & 11.14 & 19.00 & 46.77 & 2.97 & 17.78 & 35.57 & 120.10 & 1.14 & 14.02 & 28.53 & 76.82 \\
\midrule
FiLMNet    & Sub-Env & 3.09 & 14.57 & 22.89 & 54.91 & 23.30 & 28.19 & 34.50 & 71.83 & 0.60 & 8.43 & 17.62 & 44.99 \\
PointNet   & Sub-Env & 1.70 & 6.07 & 9.04 & 21.44 & 1.70 & 7.39 & 12.90 & 43.77 & 0.79 & 6.10 & 11.05 & 29.69 \\
\textbf{Transolver} & Sub-Env & \textbf{0.63} & \textbf{3.51} & \textbf{6.26} & \textbf{20.15} & \textbf{1.03} & \textbf{5.74} & \textbf{10.83} & \textbf{43.94} & \textbf{0.31} & \textbf{4.50} & \textbf{9.59} & \textbf{28.09} \\
GNOT       & Sub-Env & 4.14 & 28.44 & 47.11 & 123.93 & 4.45 & 25.14 & 40.69 & 81.89 & 1.84 & 15.97 & 29.50 & 76.21 \\
GraphUNet  & Sub-Env & 2.14 & 9.63 & 16.64 & 46.59 & 2.19 & 12.77 & 28.52 & 98.66 & 0.88 & 10.54 & 23.00 & 72.73 \\
\bottomrule
\end{tabular}
\end{adjustbox}
\end{table}

\begin{figure}[htbp]
    \centering
    \includegraphics[width=\linewidth]{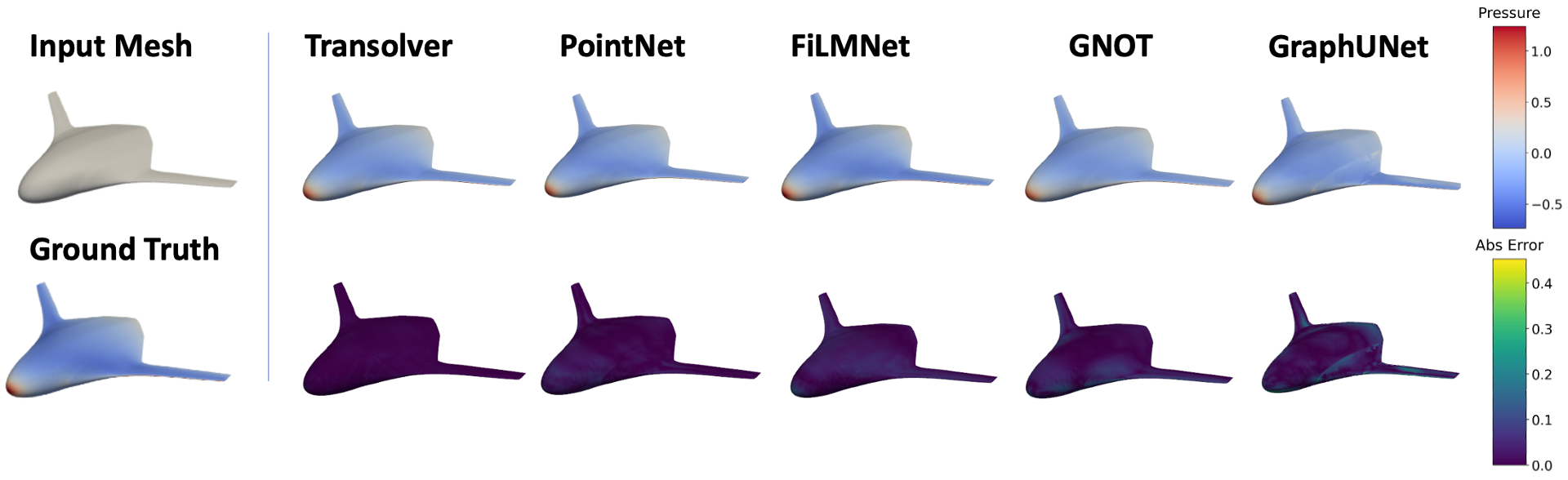}
    \caption{Qualitative field validation for the \textbf{Transolver} model trained on the \textbf{Sub-Envelope} (Group 3 UAS) regime. Each column represents a different model: the top row displays the predicted $C_p$ distribution, and the bottom row displays the absolute error ($|C_{p, \text{pred}} - C_{p, \text{CFD}}|$) compared to the RANS ground truth.}
    \label{fig:residuals_analysis}
\end{figure}

The \textbf{Winning Model} is the \textbf{Sub-Envelope Transolver}, which achieves the lowest absolute errors across all metrics, notably a $P_{99}$ pressure error of 20.15. Its performance in the friction fields ($C_{f_x} P_{95} = 10.83 \times 10^{-4}$) is particularly impressive, suggesting that its physics attention mechanism effectively captures the geometric curvature necessary to resolve viscous effects at lower Reynolds numbers. The high accuracy of these predictions is qualitatively confirmed in Fig.~\ref{fig:residuals_analysis}, which visualizes the pressure residuals for the Transolver model trained on the sub-envelope regime.Its also worth noting the "indifference to scale" observed in FiLMNet is its greatest strength. Its $P_{99}$ error remains stable across regimes, whereas PointNet and GNOT exhibit catastrophic failure when moved to the broader Full-Envelope distribution. 
A key takeaway from these results is that while \textbf{Transolver} trained on the sub-envelope achieves the highest accuracy, \textbf{FiLMNet} is the most robust and computationally efficient general architecture.

\subsection{Inverse Design }
\label{sec:results-inverse}

Inverse design is posed as conditional generation of planform parameters $\mathbf{p}\in\mathbb{R}^9$ (Table~\ref{tab:param_bounds}) given a operating condition $\boldsymbol{\mu}$ and a target lift-to-drag ratio ($L/D$). For each of the 2{,}500 test cases, we generate $100$ samples with a conditional diffusion model (CDM). As a baseline, we run multi-start gradient-based optimization (``Opt'') on the same evaluation surrogate, with $100$ random initial seeds per condition and projected updates (500 steps using Adam) to respect bounds. A hybrid (``CDM$\rightarrow$Opt'') refines each CDM sample via a short local optimization pass (200 steps using Adam).

We report two aspects per condition, then average across all conditions: (i) \emph{accuracy} w.r.t.\ the $L/D$ target using $R^2$ (higher is better), RMSE (lower), and MAE (lower); and (ii) \emph{diversity} of generated parameters using mean pairwise distance (MPD; larger spread is better) and mean nearest-neighbor distance (MinDist; larger indicates more even spacing) computed in standardized parameter space.

\begin{table}[ht]
\centering
\caption{Inverse design results averaged over $2{,}500$ conditions. Accuracy (computed w.r.t.\ surrogate-predicted $L/D$) and diversity; runtime is wall-clock per condition to produce 100 samples. Better: $R^2\uparrow$, RMSE/MAE$\downarrow$, MPD/MinDist$\uparrow$.}
\label{tab:inverse-results}
\scriptsize
\setlength{\tabcolsep}{3pt}
\renewcommand{\arraystretch}{1.15}

\resizebox{\textwidth}{!}{%
\begin{tabular}{l c c c c c c}
\toprule
\multirow{2}{*}{\textbf{Method}} & \multicolumn{3}{c}{\textbf{Accuracy to target $L/D$}} & \multicolumn{2}{c}{\textbf{Diversity in parameter space}} & \textbf{Runtime} \\
\cmidrule(lr){2-4}\cmidrule(lr){5-6}\cmidrule(l){7-7}
& $R^2\uparrow$ & RMSE$\downarrow$ & MAE$\downarrow$
& \shortstack{MPD\\(spread)$\uparrow$}
& \shortstack{MinDist\\(uniformity)$\uparrow$}
& \shortstack{Time/cond\\(100 samples)} \\
\midrule
CDM &
0.99601 &
\shortstack{$0.401$\\$\pm 0.170$} &
\shortstack{$0.330$\\$\pm 0.162$} &
\shortstack{\textbf{3.976}\\$\pm$ \textbf{0.161}} &
\shortstack{\textbf{1.934}\\$\pm$ \textbf{0.115}} &
0.3 s \\
Opt &
\textbf{0.99998} &
\shortstack{\textbf{0.002}\\$\pm$ \textbf{0.032}} &
\shortstack{\textbf{0.002}\\$\pm$ \textbf{0.031}} &
\shortstack{$3.328$\\$\pm 0.450$} &
\shortstack{$1.351$\\$\pm 0.278$} &
3.0 s \\
CDM$\to$Opt &
0.99811 &
\shortstack{$0.227$\\$\pm 0.196$} &
\shortstack{$0.179$\\$\pm 0.166$} &
\shortstack{$3.899$\\$\pm 0.225$} &
\shortstack{$1.853$\\$\pm 0.147$} &
0.9 s \\
\bottomrule
\end{tabular}%
}

\vspace{0.35em}
\footnotesize
Runtime notes: Opt uses 500 PGD steps with Adam; CDM is sampling only; CDM$\to$Opt is CDM sampling + 200 PGD steps with Adam. Runtimes measured on a single NVIDIA RTX 4090 GPU.
\end{table}

As shown in Table~\ref{tab:inverse-results}, pure multi-start optimization (Opt) achieves nearly perfect accuracy with respect to the prescribed $L/D$ targets ($R^2 \approx 1.0$, near-zero MAE), but converges to a narrower set of solutions, resulting in the lowest diversity (lowest MPD and MinDist). In contrast, the conditional diffusion model (CDM) acts as a strong generative prior that explores a much broader and more uniform region of the valid design space (highest MPD and MinDist), though its zero-shot samples are slightly less precise in hitting the exact target. 

The hybrid CDM$\rightarrow$Opt approach offers a highly effective middle ground: a short 200-step refinement improves the target-tracking accuracy of the CDM samples (reducing MAE by nearly half) while largely preserving the rich diversity of the generative prior, resulting in significantly higher geometric variety than random-seed optimization. \textit{In terms of speed, CDM sampling is $\sim$10$\times$ faster than Opt per condition (about $0.3$\,s vs.\ $3$\,s for 100 samples).} 

\subsubsection*{CFD validation of inverse designs}
To verify that inverse designs remain valid under high-fidelity flow solving, we re-simulated (with FUN3D) one \textit{CDM$\rightarrow$Opt} sample per test condition from the same 2,500-case test set utilized for our field-resolved surrogate benchmarking. We then compared the FUN3D-evaluated $L/D$ against the target $L/D$ for those designs. Across all conditions, correlation is excellent ($R^2=0.998$), supporting the functionality of the diffusion–optimization pipeline.

\begin{figure}[h!]
  \centering
  \includegraphics[width=0.6\linewidth]{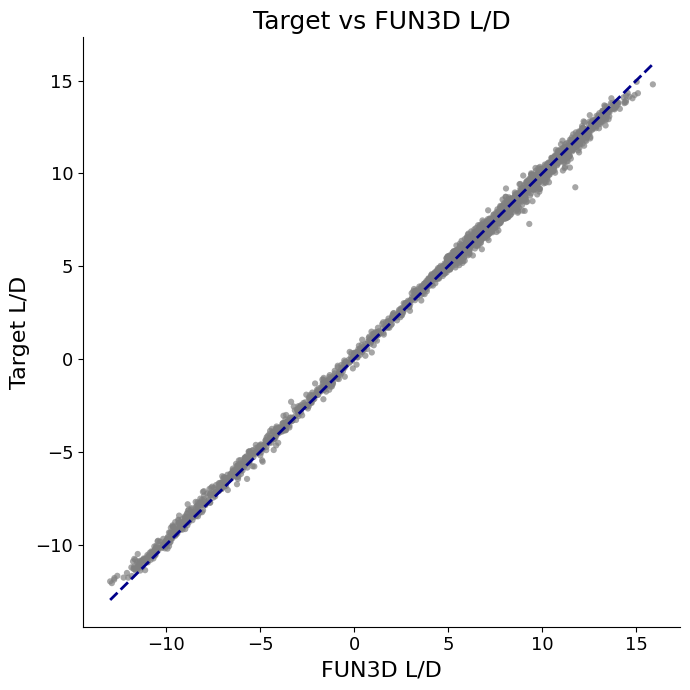}
  \caption{CFD-validated inverse designs: FUN3D $L/D$ versus surrogate-predicted $L/D$ for one \textit{CDM$\rightarrow$Opt} sample per condition. Overall correlation is $R^2=0.998$.}
  \label{fig:cfd_ld_validation}
\end{figure}

\section{Conclusion, Limitations, and Future Work}
We have introduced a data-driven framework for the aerodynamic design of Blended Wing Body aircraft, supported by \textbf{BlendedNet++}, the largest publicly available dataset of field-resolved BWB RANS simulations. By benchmarking five classes of deep learning surrogates, we identified architectures capable of predicting surface pressures and skin friction with high fidelity, enabling rapid analysis without repeated CFD solves.

Furthermore, we demonstrated that generative artificial intelligence can fundamentally accelerate the design process. Our hybrid inverse design strategy which couples conditional diffusion for global exploration with gradient-based refinement to successfully generate valid BWB planforms that meet its aerodynamic efficiency targets. Validation against high-fidelity CFD confirms the physical realizability of these designs.

\paragraph{Limitations and design choices.}\mbox{}\\
(1) \emph{Single condition per geometry (by design).} BlendedNet++ evaluates each geometry at a single flight condition, which limits conclusions about off-design behavior (e.g., stall onset, polars, or broad operating envelopes). We make this choice intentionally to concentrate CFD budget on maximizing \emph{geometric coverage}, enabling rigorous \emph{geometry-disjoint} generalization tests. In this framing, the primary task is \emph{geometric generalization} (how performance and fields vary with shape at a fixed operating point distribution), not \emph{operational generalization} (how aerodynamics vary across conditions for a fixed shape). A natural extension is to introduce \emph{multi-condition cohorts} (the same geometry evaluated at multiple conditions) to jointly study geometry $\times$ condition generalization.

\noindent(2) \emph{Surface fields only.} We focus on $C_p$ and $\mathbf{C}_f$ over the surface to keep storage and I/O manageable and to target the quantities most directly used for forces and moments. This precludes operator-learning studies on full volumetric flow fields in the current release. 

\noindent(3) \emph{CFD and meshing fidelity.} Automated meshing at scale can introduce variability in near-wall resolution and element quality, and steady RANS with a single turbulence model may deviate from true flow physics for separated or highly three-dimensional regimes. These choices reflect a balance between throughput and fidelity for a first large release. 

\noindent(4) \emph{Subsonic regime and UAS focus.} Regarding the flight envelope, BlendedNet++ is intentionally tailored to the subsonic regime ($M < 0.5$). This choice is motivated by the operational requirements of Group 3 tactical UAS. While transonic extensions would introduce challenging shock-wave physics and wave drag phenomena, such conditions fall outside the intended tactical mission profiles of the airframes studied here. Future iterations focusing on commercial transport-scale BWBs would require an expansion into the transonic regime, necessitating shock-capturing solvers and refined leading-edge meshing.

\paragraph{Future directions.}
We see several extensions as especially impactful: (i) matched \emph{multi-condition} cohorts per geometry (and potentially unsteady/transient cases) to probe condition coupling (ii) inclusion of \emph{volumetric} fields for visualizing 3D streamlines and vorticity which allows designers to identify the origin of spanwise flow and tip vortices that contribute to induced drag. (iii) curated \emph{higher-fidelity} and/or experimental subsets (e.g., DES/LES or wind-tunnel comparisons) for cross-validation (iv) richer, \emph{constrained} inverse-design tasks that couple aerodynamics with structural and stability constraints which incorporates internal volume requirements and $C_M$ trim targets through secondary geometric surrogates and multi-objective refinement losses and (v) community-facing \emph{leaderboards/challenge tracks} on the provided splits to encourage fair, reproducible progress.

Overall, BlendedNet++ is intended as a foundation rather than a finished endpoint: the benchmark configurations are representative baselines (not claimed optimal), and the dataset and code are structured to support systematic scaling, ablations, and community-driven extensions.


\section*{Declaration of competing interest}
The authors declare that they have no known competing financial interests or personal
relationships that could have appeared to influence the work reported in this paper.

\section*{Declaration of generative artificial intelligence and artificial intelligence-assisted technologies in the manuscript preparation process}
During the preparation of this work, the author(s) used ChatGPT (OpenAI) to assist with language editing and organization of the manuscript. After using this tool, the author(s) reviewed and edited the content as needed and take full responsibility for the content of the published article.

\section*{Acknowledgments}

DISTRIBUTION STATEMENT A. Approved for public release. Distribution is unlimited.





\section*{Code availability}
Code to reproduce preprocessing, training, and evaluation will be publicly released in a permanent repository upon publication (an anonymized link can be provided during review). 
A link will be added in the final version.

\bibliography{sample} 

\newpage
\appendix
\section*{Appendix}

\subsection{Supplementary Dataset Distributions}
\begin{figure}[H] 
  \centering
  \includegraphics[width=0.9\linewidth]{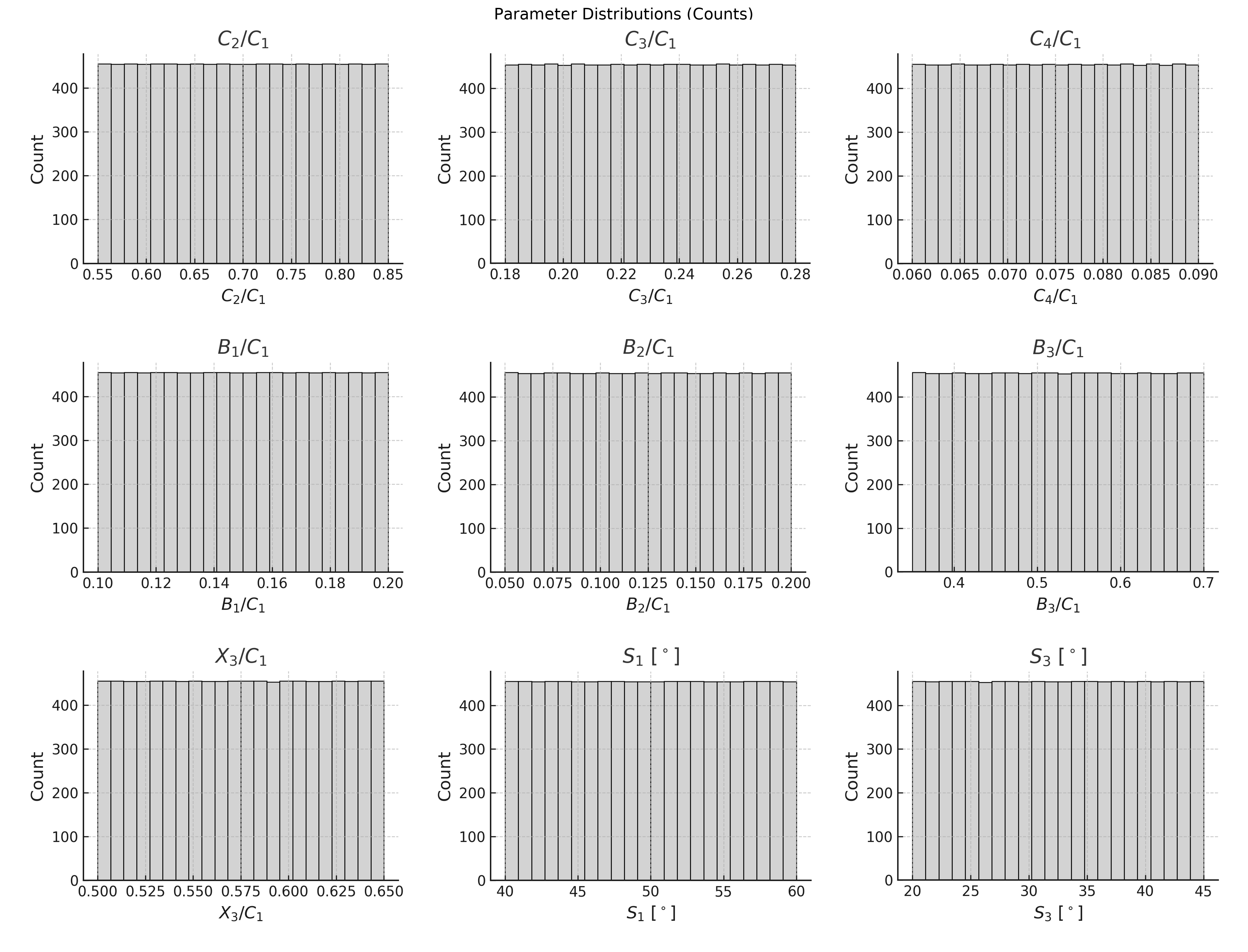}
  \caption{Distributions (counts) of normalized planform parameters for BlendedNet++. Each panel shows the marginal histogram of a single parameter within the bounds listed in Table~\ref{tab:param_bounds}.}
  \label{fig:geom_param_hists}
\end{figure}

\begin{figure}[H]
    \centering
    \includegraphics[width=\linewidth]{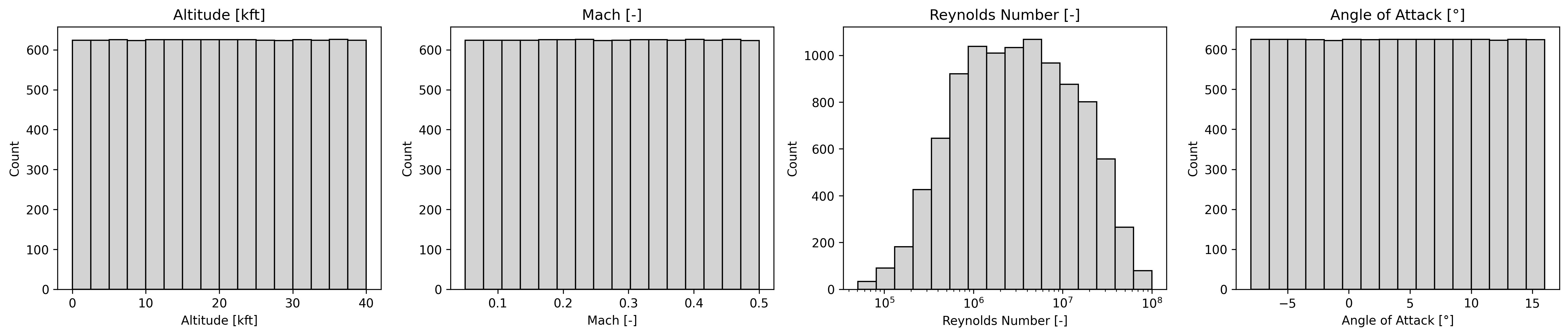}
    \caption{Flight-condition distributions for BlendedNet++: altitude, Mach number, Reynolds number (log-scaled on the horizontal axis), and angle of attack.}
    \label{fig:fc_hist}
\end{figure}

Figures~\ref{fig:geom_param_hists} and~\ref{fig:fc_hist} report the distributions of the sampled variables. Figure~\ref{fig:geom_param_hists} shows histograms of the normalized planform parameters within the bounds of Table~\ref{tab:param_bounds}. Figure~\ref{fig:fc_hist} shows the flight-condition marginals (altitude, Mach, $\log_{10}Re_L$, and $\alpha$).

\subsection{BlendedNet vs.\ BlendedNet++ Comparison}

\begin{figure}[H]
  \centering
  \includegraphics[width=0.45\linewidth]{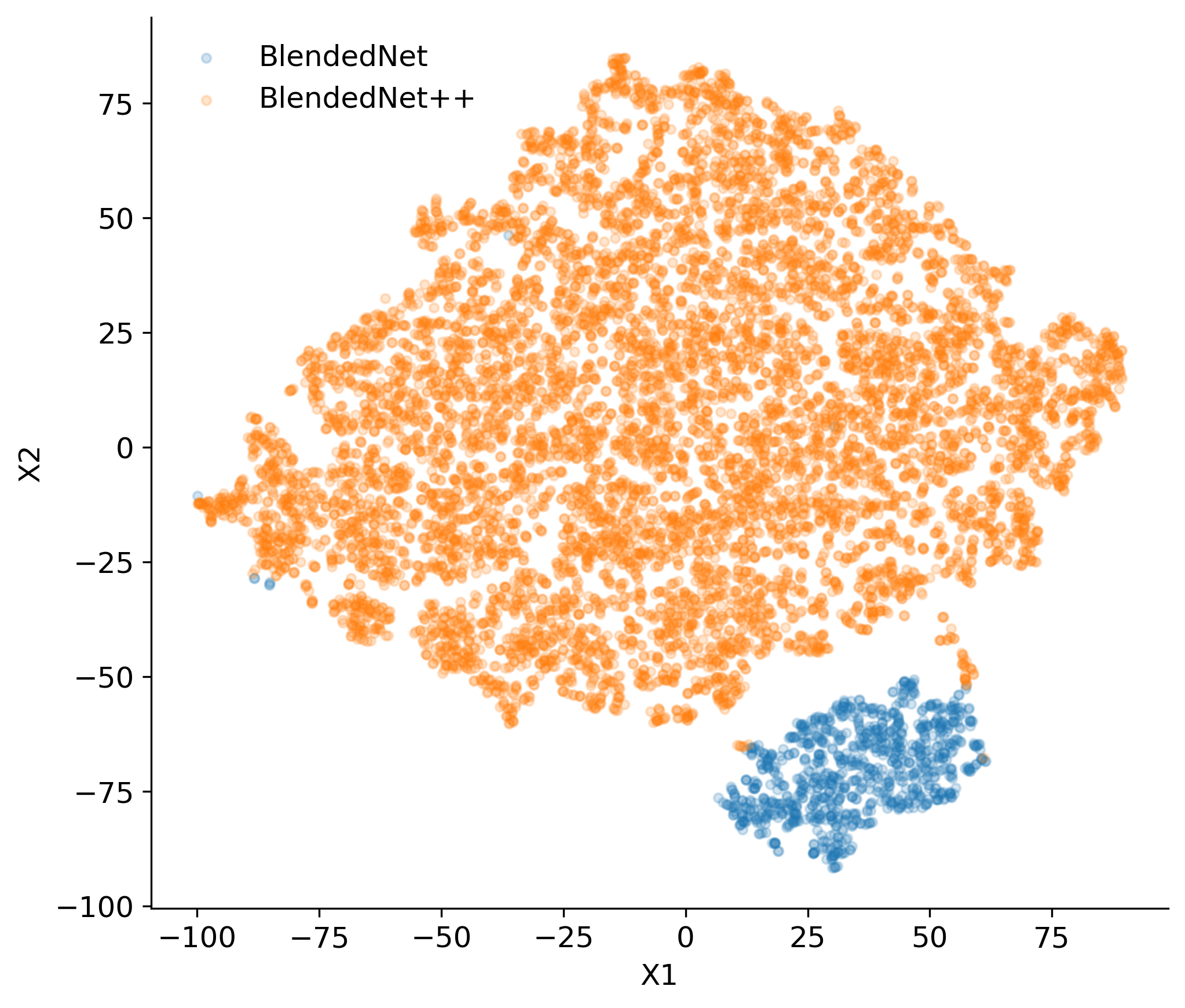}
  \caption{t-SNE of PointNet autoencoder latent space for BlendedNet and BlendedNet++. Each point is one geometry.}
  \label{fig:app_tsne_pointnet_ae}
\end{figure}

\noindent A t-SNE embedding of the latent space learned by a PointNet autoencoder, trained on both datasets, is shown in Figure~\ref{fig:app_tsne_pointnet_ae}, where each point corresponds to a geometry and is colored by dataset.

\begin{figure}[H]
  \centering
  \includegraphics[width=0.6\linewidth]{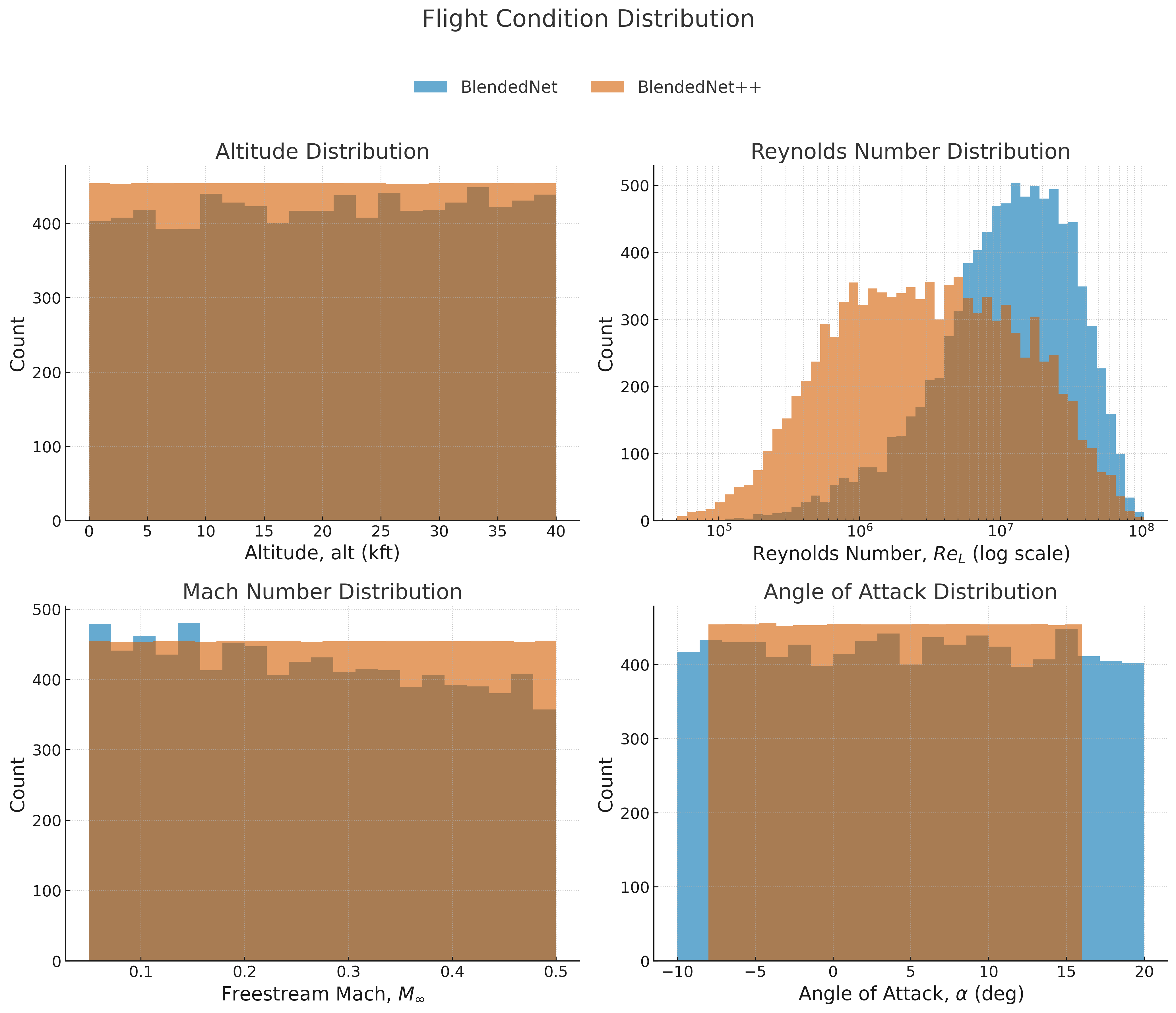} 
  \caption{Distribution of flight conditions using Altitude, Reynolds Number, Mach Number and Angle of Attack.}
  \label{fig:app_flight}
\end{figure}

\noindent We also visualize the flight-condition distribution to show sampling coverage across Altitude, Reynolds Number, Mach Number and Angle of Attack (Figure~\ref{fig:app_flight}).

\begin{figure}[H]
  \centering
  \includegraphics[width=0.6\linewidth]{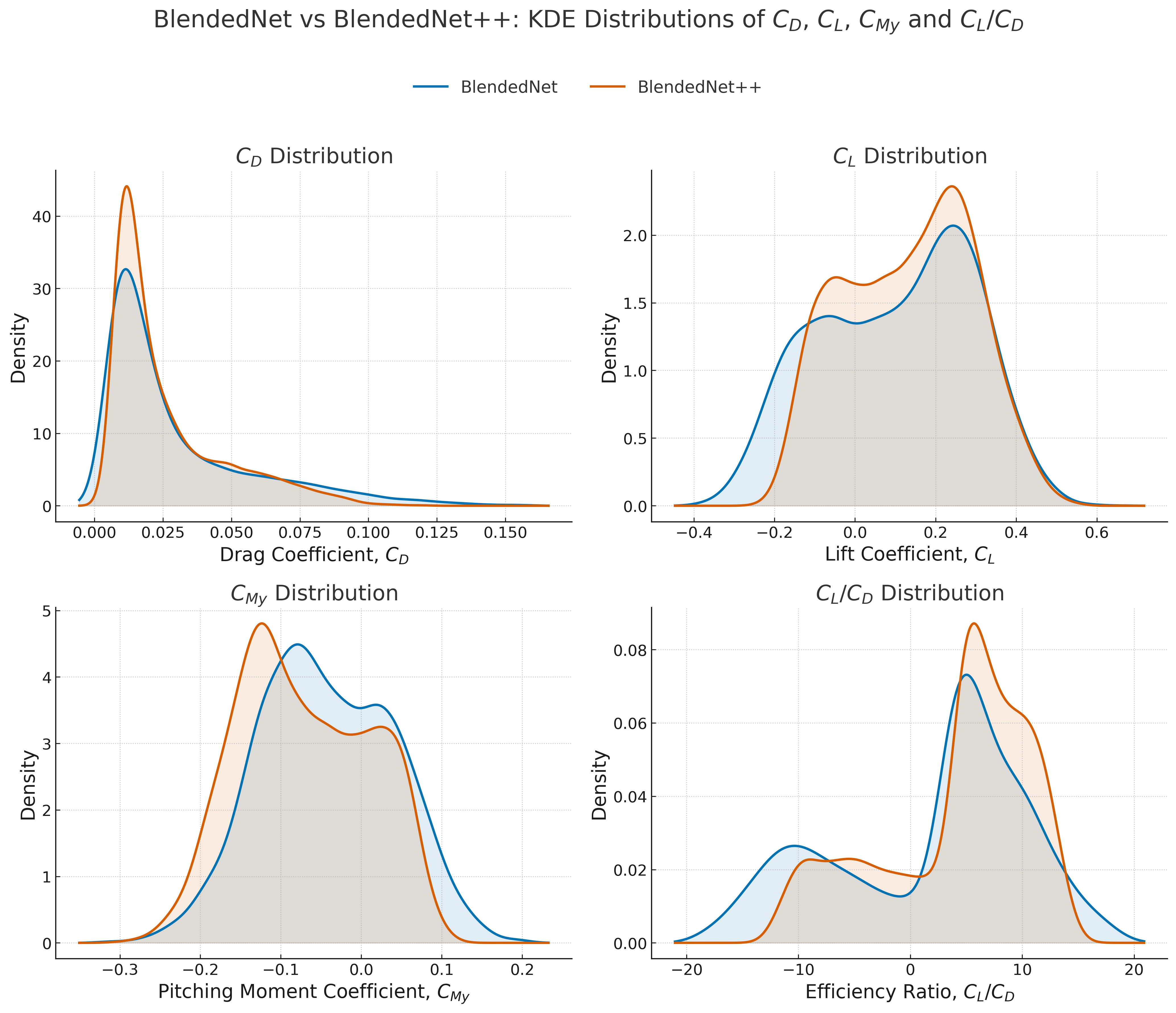}
  \caption{Distributions of $C_D$, $C_L$, $C_{My}$, and efficiency $C_L/C_D$ for BlendedNet vs.\ BlendedNet++.}
  \label{fig:app_coeff_hists}
\end{figure}

\noindent Finally, Figure~\ref{fig:app_coeff_hists} compares the distributions of the aerodynamic coefficients $C_D$, $C_L$, $C_{M}$, and the efficiency ratio $C_L/C_D$.

\subsection{Inverse Design: Optimization \& Diffusion Details}

\subsubsection{Conditional diffusion configuration}
We train a conditional denoising model that maps a noisy geometry vector to new planform parameters. The denoiser is a residual multilayer perceptron architecture with a hidden width of 512 and a depth of 6 residual blocks. Each block incorporates \textsc{SiLU} activations and Layer Normalization. To avoid notation ambiguity with the angle of attack ($\alpha$), we denote the diffusion coefficients as $a_t$. The diffusion process uses $T=1000$ timesteps with a cosine noise schedule and sinusoidal timestep embeddings.

The model is trained for $10^5$ steps using the AdamW optimizer with a learning rate of $2 \times 10^{-4}$ and a weight decay of $1 \times 10^{-4}$. Training was performed using a batch size of 128 with Automatic Mixed Precision (AMP) and a gradient clipping threshold of 1.0. The conditioning vector $\boldsymbol{\mu}$ comprises five features: altitude [kft], $\log_{10} Re_L$, $M_\infty$, $\alpha$ [deg], and the target $L/D$ ratio. 

The normalization pipeline for the 9-dimensional geometry vector $\mathbf{p}$ (excluding the fixed $C_1 = 1000$ m) involves a symmetric min-max scaling to $[-1,1]$ followed by a quantile transformation fitted with a normal output distribution. At inference, we draw $K=100$ independent Gaussian noise seeds per condition and execute the full reverse sampling process to generate candidate geometries.

\subsubsection{Surrogate Objective Function}
Both optimization and refinement tasks utilize a differentiable surrogate model trained to predict $L/D$. The surrogate employs a residual regression architecture with a hidden width of 256, a depth of 3 blocks, and a dropout rate of 0.2. It was trained for 120 epochs using AdamW ($lr=1 \times 10^{-3}$, weight decay $1 \times 10^{-5}$) to minimize Mean Squared Error (MSE).

\subsubsection{Projected Gradient Descent (PGD)}
We implement a batched PGD routine that operates directly on the GPU to achieve significantly higher throughput than sequential processing. The objective is the squared error $(L/D_{\text{pred}} - L/D_{\text{target}})^2$. Updates are computed in the standardized latent space using the Adam optimizer with a learning rate of 0.05. After each update, the parameters are projected back into the physical feasible box $\mathcal{B}$ by inverting the normalization, clipping values to the bounds defined in Table~\ref{tab:param_bounds}, and re-applying the forward normalization.

Two optimization regimes are established:
\begin{enumerate}
    \item \textbf{Multi-start Baseline (Opt):} 100 random seeds are sampled uniformly from the parameter bounds and optimized for 500 steps.
    \item \textbf{Diffusion–Optimization Hybrid (CDM$\rightarrow$Opt):} 100 samples are generated by the conditional diffusion model and locally refined for 200 steps to ensure exact adherence to the target $L/D$.
\end{enumerate}
In both cases, we report the candidate that minimizes the final prediction error for each condition.

\end{document}